%% file: 1main.tex
\pdfoutput=1

\documentclass[11pt]{article}

\usepackage{ACL2023}

\usepackage{times}
\usepackage{latexsym}
\usepackage[T1]{fontenc}
\usepackage[utf8]{inputenc}
\usepackage{microtype} 
\usepackage{inconsolata}
\usepackage{float}
\usepackage{graphicx}
\usepackage{booktabs}
\usepackage{makecell}
\usepackage{enumitem}
\usepackage{caption}
\usepackage{subcaption}
\usepackage[many]{tcolorbox}
\usepackage{adjustbox}

\usepackage{multirow} 
\usepackage{longtable}
\usepackage{siunitx}
\usepackage{booktabs}
\usepackage{makecell}

\makeatletter
\newcommand\notsosmall{\@setfontsize\notsosmall\@xpt\@xiipt}
\makeatother

\newtcolorbox{boxI}{
    colback = sub, 
    colframe = main, 
    boxrule = 0pt, 
    toprule = 3pt
}
\definecolor{main}{HTML}{5989cf}
\definecolor{sub}{HTML}{cde4ff}

\title{Revisiting Word Embeddings in the LLM Era}

\author{
  Yash Mahajan$^1$ \quad Matthew Freestone$^1$ \quad Naman Bansal$^1$ \\ \quad \textbf{Sathyanarayanan Aakur}$^1$ \quad \textbf{Santu Karmaker}$^2$ \\
  $^1$Auburn University \quad $^2$University of Central Florida \\
  \texttt{\{yzm0034, maf0083,nzb0040, san0028\}@auburn.edu, santu@ucf.edu}
}
\begin{document}
\maketitle

\begin{abstract}
%Learning meaningful word embeddings is key to training a robust language model. 

Large Language Models (LLMs) have recently shown remarkable advancement in various NLP tasks. As such, a popular trend has emerged lately where NLP researchers extract word/sentence/document embeddings from these large decoder-only models and use them for various inference tasks with promising results. However, it is still unclear whether the performance improvement of LLM-induced embeddings is merely because of scale or whether underlying embeddings they produce significantly differ from classical encoding models like Word2Vec, GloVe, Sentence-BERT (SBERT) or Universal Sentence Encoder (USE). This is the central question we investigate in the paper by systematically comparing classical decontextualized and contextualized word embeddings with the same for LLM-induced embeddings. Our results show that LLMs cluster semantically related words more tightly and perform better on analogy tasks in decontextualized settings. However, in contextualized settings, classical models like SimCSE often outperform LLMs in sentence-level similarity assessment tasks, highlighting their continued relevance for fine-grained semantics.

\end{abstract}

\input{2introduction}

\input{3Related_Work}

\input{4.0_Decontexualized}

\input{4.1_Contextualized}

\input{5results}

\input{6conclusion}
\input{7Limitation}

\bibliography{custom,santu}
\bibliographystyle{acl_natbib}

\input{8appendix}

% FUTURE WORK
% Comparing different layers might have already been done (ELMo, BERT, GPT-2):
% https://aclanthology.org/D19-1006.pdf 
% “For one, we find that the contextualized representations of all words are not isotropic in any layer of the contextualizing model. While representations of the same word in different contexts still have a greater cosine similarity than those of two different words, this self-similarity is much lower in upper layers.”
% Central Kernel Alignment: 
% https://arxiv.org/abs/1905.00414 
% We introduce a similarity index that measures the relationship between representational similarity matrices and does not suffer from this limitation

% Small Changes
% We could do a rewrite with more direct conclusions and more methodology. That was a big thing in reviews, but we resolved it in rebuttal, I think.

% Other work
% Binary embeddings with EC proposal with Tauritz; if promising, I can flesh that out.
% https://www.overleaf.com/project/65d50cde71dd06197f90957a 

\end{document}

%% file: 2introduction.tex
\section{Introduction}

% {The field of NLP has witnessed a remarkable evolution in text representation methodologies over the past decade. 
% % This transformation began with the introduction of
% Starting with 

% and semantic analysis for deep NLP research. These models laid the foundation for representing words as dense vectors, capturing semantic relationships, and enabling machines to process language more effectively. The complexity and scale of embedding models have since increased dramatically, marked by the advent of.

Word2Vec~\cite{mikolov2013efficient} and GLoVe~\cite{pennington-etal-2014-glove}, which revolutionized The field of NLP and word embedding techniques by representing words as dense vectors. The complexity and scale of embedding models have since increased dramatically.
Transformer-based architecture like BERT-based models~\cite{devlin2018bert}, RoBERTa~\cite{liu2019roberta} 
expanded language representation capabilities by providing context-aware embeddings for words and longer sequences. The most recent paradigm shift came with Large Language Models (LLMs) like
% etc. These models expanded the capabilities of language representation by providing context-aware embeddings for words and longer sequences. The most recent paradigm shift in NLP came with the emergence of Large Language Models (LLMs) like 
GPT~\cite{brown2020language}, PaLM~\cite{chowdhery2022palm}, 
LLaMA~\cite{touvron2023llama}, etc. A popular trend has emerged where NLP researchers extract word/sentence/document embeddings from these large decoder-only models for various inference tasks with promising results. However, it is still unclear whether the performance improvement of LLM-induced embeddings is merely because of scale or whether the underlying embeddings they produce significantly differ from classical models.

%Table~\ref{tab:samples} showcases examples generated for each task using the anchor words.

To explore this, we conducted an in-depth investigation of word embedding similarity in two settings: 1) decontextualized and 2) contextualized for both classical models and LLMs. In the decontextualized setting, we generated embeddings for $\approx 80,000$ words, {with curtailed datasets for pretrained Word2Vec ($\approx 50K$) and GloVe ($\approx 60K$) due to vocabulary limitations. We analyzed them using word-pair similarity and word analogy tasks.} 
% \yash{It's worth noting that for static embedding models like Word2Vec and GloVe, which have finite vocabulary sizes, we used a curtailed dataset of approximately 50,000 and 60,000 recognized words respectively, to ensure a fair comparison}. 
For the contextualized setting, we selected \textit{anchor words} (verbs, nouns, or adjectives) and created multiple sentences using them to provide context. We then extracted the embeddings of these anchor words for evaluation. More specifically, we examined embedding similarity across nine diverse variational tasks, including \textit{synonym}, \textit{antonym}, \textit{negation}, \textit{jumbling}, \textit{paraphrase}, \textit{questionnaire}, \textit{exclamation}, and \textit{polysemy}. To compare the models in contextualized settings, we performed three distinct similarity analyses: 1. \textit{Anchor Inter-Contextual Variance}: measuring the variance of an anchor word embedding across different contexts; 2) \textit{Anchor Contextual Deviation}: Assessing how context influences anchor word embeddings compared to their decontextualized counterparts; 3) \textit{Sentence Similarity}: Measuring a model's ability to capture linguistic variations at a sentence level.

Our results show that LLMs cluster semantically related words more tightly and perform better on analogy tasks in decontextualized settings. However, in contextualized settings, classical models like SimCSE outperform LLMs in sentence-level tasks, highlighting their continued relevance.

%% file: 3Related_Work.tex
\vspace{-1.5mm}
\section{Related Work}
\vspace{-1.5mm}

\iffalse
cite this:
https://aclanthology.org/2021.acl-long.65.pdf
    - demonstrate the issue in machine translation; misinterpreting pronoun and polysemous words during machine translation. 

\fi

Text representation is a fundamental pursuit in NLP research, and we have witnessed a remarkable evolution in text representation methodologies over the past decade. This transformation can be grouped into four generations: 1) Classic Decontexualized Word Embeddings like Word2Vec~\cite{mikolov2013efficient} and GloVe~\cite{pennington-etal-2014-glove}; 2) Transformer-based contextualized Embeddings like BERT~\cite{devlin2018bert}, BART~\cite{lewis2019bart}, and RoBERTa~\cite{liu2019roberta}; 3) Sentence Encoders such as LASER~\cite{Artetxe_2019}, Universal Sentence Encoder (USE)~\cite{cer2018universal}, and SentenceBERT (SBERT)~\cite{reimers2019sentencebert}; and 4) Large Language Model (LLM) induced embeddings like GPT~\cite{brown2020language}, PaLM~\cite{chowdhery2022palm}, LLaMA~\cite{touvron2023llama}, OpenELM~\cite{openelm}, OLMo~\cite{olmo} etc.

% began with the introduction of , which revolutionized word embedding techniques and semantic analysis in deep NLP research. 

% These models laid the foundation for representing words as dense vectors, capturing semantic relationships and enabling machines to process language more effectively. Following these breakthroughs, the complexity and scale of embedding models increased dramatically. The introduction of transformer-based architectures marked a significant leap forward. Models like BERT~\cite{devlin2018bert}, BART~\cite{lewis2019bart}, and RoBERTa~\cite{liu2019roberta} expanded the capabilities of language representation by providing context-aware embeddings for words and longer sequences. This advancement was further refined with the development of sentence encoders such as LASER~\cite{Artetxe_2019}, Universal Sentence Encoder (USE)~\cite{cer2018universal}, and SentenceBERT (SBERT)~\cite{reimers2019sentencebert}, which focused on capturing meaning at the sentence level. The most recent paradigm shift in NLP came with the emergence of Large Language Models (LLMs) like GPT~\cite{brown2020language}, PaLM~\cite{chowdhery2022palm}, LLaMA~\cite{touvron2023llama}, OpenELM~\cite{openelm}, OLMo~\cite{olmo} etc. 

Previous work by~\citet{haber2021patterns,fournier2020analogies,haber2024polysemy, ethayarajh2019contextual,yash,souvika-USE} have investigated how transformer-based models capture word context to varying degrees. In contrast, previous work by \citet{peters2018dissecting,li2024probing,miaschi2020contextual} has focused on extracting context-independent word representations for tasks such as word analogy.

Recent LLMs, with their unprecedented scale and capabilities, have demonstrated remarkable success across various NLP tasks ~\cite{bubeck2023sparks,dai2022can,du2022glam,smith2022using,zero-shot-souvika,Bangla-Word-Analogy}. This has motivated multiple NLP researchers to extract word/sentence embeddings from these decoder-only models and use them for other downstream tasks different from text generation~\cite{jiang2023scaling,an2024capturing}. Despite these advancements, the fundamental medium of written language has remained constant. While the similarity and relatedness of words have not inherently changed, the models' approach to treating words and their similarities has evolved significantly. This raises important questions about the nature of embeddings generated by LLMs compared to those created by traditional encoding models like Word2Vec or Sentence-BERT. Indeed, little is known about the fundamental nature of these LLM-induced embeddings and how they differ from classical embeddings. It is also unclear how these word embeddings differ from each other in both contextualized and decontextualized settings.

%% file: 4.0_Decontexualized.tex
\vspace{-2mm}
\section{Comparing Decontextualized Embeddings: LLM vs. Classical}
\vspace{-1.5mm}
We conduct a comparative study of two groups of models: 1) Large Language Models (LLMs) (decoder models with over 1B parameters) and 2) ``Classical'' (models with under 1B parameters) in terms of their decontextualized word embeddings. To be more specific, we selected thirteen models for our analysis, including seven LLMs and six classical models. The LLMs include: LLaMA2-7B and LLaMA3-8B (both dim $=4096$) from Meta AI \cite{touvron2023llama}, OpenAI's embedding model ADA-002 (dim $=1536$), and Google's PaLM2 embedding model Gecko-001 (dim $=768$)~\cite{anil2023palm}, OLMo-8B (dim $=4096$) ~\cite{olmo}, OpenELM-3B (dim $=3072$)~\cite{openelm} and, Mistral-8B (dim $=4096$)~\cite{mistral}. To more clearly see the differences between these models and older (``classical'') ones, Meta AI's LASER (dim $=1024$)~\cite{Artetxe_2019}, Universal Sentence Encoder (USE) (dim $=512$)~\cite{cer2018universal}, SimCSE (dim $=1024$) ~\cite{simcse}, SBERT (dim=384)~\cite{reimers2019sentencebert}, Word2vec (dim=300)~\cite{mikolov2013efficient} and GloVe (dim=300)~\cite{pennington-etal-2014-glove}.

Decontextualized embeddings are obtained by inputting single words into each model's tokenizer. For models using single-token inputs, we utilize the final hidden state. For subword tokenization, we average the final hidden states of the tokens. Using these decontextualized embeddings,  we conduct the following three comparative analyses. %of the decontextualized embeddings.

\begin{itemize}[leftmargin=*,itemsep=0ex,partopsep=-0.5ex,parsep=-0.5ex]
    \item {\textit{Word-Pair Similarity Comparison}}%:  We compare the cosine similarity distributions of $\approx80,000$ word pairs spread across three categories: \textit{Semantically Related, Morphologically Related, and Unrelated words}. This analysis helps us compare how different embedding techniques capture various types of word relationships.
    
    \item {\textit{Analogy Task Based Comparison}}%: For a given word analogy task of the form $a:b::c:d$ (``a is to b as c is to d''), we compare the accuracies of various embedding techniques using vector arithmetic in the latent vector space. This analysis helps us compare the compositional power of different embedding techniques. 
    
    %representations of $a$, $b$, and $c$ to make the result the nearest neighbor of $d$ are measured.

    \item {\textit{Similarity Correlation Analysis}}
    % \vspace{-4mm}
\end{itemize}
\vspace{-2mm}

%{new_plots/cosine_distribution_violinplot_transposed.png}
\begin{figure}[!htb]
    \centering
    \includegraphics[width=\linewidth, height=8cm]{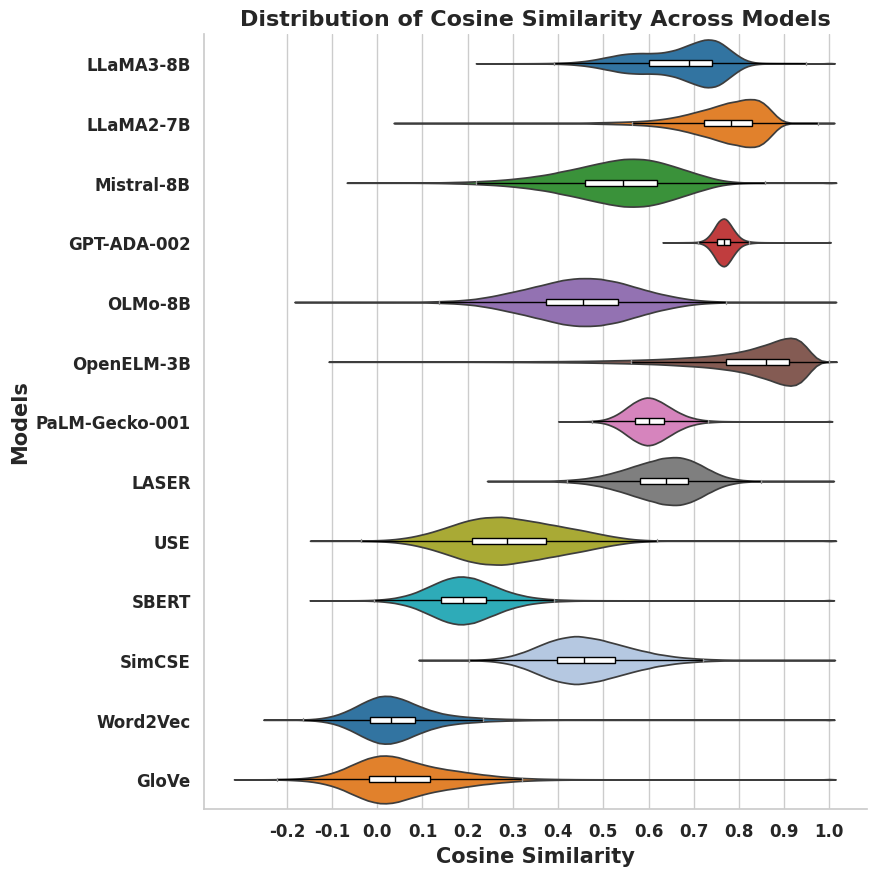}
    \vspace{-6mm}
    \caption{The distribution of cosine similarities between all pairs of words for each model.}
    \label{fig:sim-all-words}
    \vspace{-4mm}
\end{figure}

\begin{figure*}[!bht]\small
    \centering
    \includegraphics[width=0.8\linewidth, height=9.2cm]{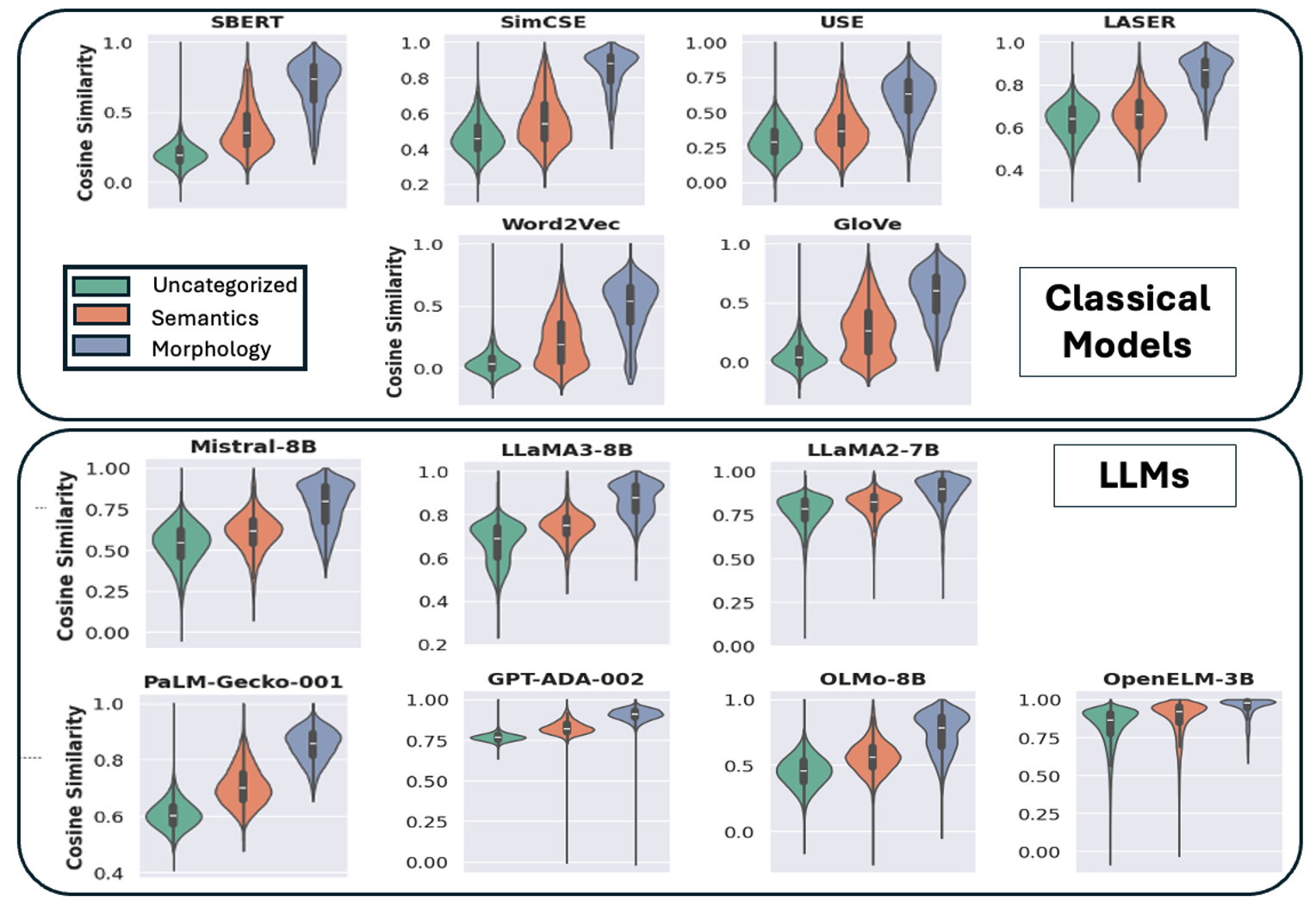}
    \vspace{-3mm}
    \caption{Violin box plot showing the distribution of cosine similarities for random, morphologically related, and semantically related pairs of words for each model.}
    \vspace{-3mm}
    \label{fig:sim-and-dissim-words}
\end{figure*}

\subsection{Word-Pair Similarity Comparison} \label{similar-closeness}

\textbf{RQ-1:} \textit{How do LLM-induced decontextualized embeddings differ from classical ones in terms of the expected cosine similarity for a randomly chosen pair of words?}

We sampled a corpus of approximately $80,000$ distinct words from WordNet~\cite{fellbaum1998wordnet} using all lemmas from all synsets and computed the cosine similarity for all $\approx 6.4$B word pairs. 
% all each pair of words using their vector representations extracted from each embedding model. The distribution of cosine similarities of all pairs of words ($\approx 6.4$B pairs) is shown in Figure \ref{fig:sim-all-words}.
The distribution for each model is meaningfully different, demonstrating significant differences in terms of their latent semantic spacing. 
% Interestingly, classical models such as SBert and USE demonstrated the lowest similarity scores (left-skewed), and LLMs like OpenELM demonstrated high similarity scores. %This demonstrates that classical models are more apt for capturing word semantics than a few LLMs

{Notably, static word embedding models and transformer-based classical models like SBert and USE showed lower similarity scores for random word pairs, resulting in left-skewed distributions. In contrast, LLMs like OpenELM exhibited higher overall similarity scores.
Due to vocabulary size limitations, Word2Vec and GloVe recognized only about $\approx 50K$ and $\approx 60K$ words, respectively. As such, we evaluated these models on a curtailed dataset containing only their recognized words.}

%These findings highlight the trade-offs between architectural choices and resulting latent space characteristics in different embedding models.

% \yash{Notably, static word embedding models exhibited an ideal distribution characterized by lower similarity scores for random word pairs. Similarly, transformer-based classical models such as SBert and USE demonstrated low similarity scores, resulting in left-skewed distributions. In contrast, Large Language Models (LLMs) like OpenELM showed higher similarity scores overall.
% An important consideration in this analysis is the vocabulary size limitation of static embedding models like Word2Vec and GloVe. Out of the 80,000 words in our corpus, Word2Vec and GloVe recognized approximately 50,000 and 60,000 words, respectively. To ensure a fair comparison while acknowledging these inherent constraints, we evaluated these two models on a curtailed dataset containing only the words they recognized.
% These findings provide valuable insights into the semantic representations of different embedding models, highlighting the trade-offs between specific architectural choices and the resulting latent space characteristics.
% }

\vspace{-2mm}
\begin{boxI}\small
\textit{\textbf{Finding-1}: LLM-induced embeddings, particularly OpenELM, ADA, and LLaMA, yield higher expected cosine similarity for a random pair of words than the same for PaLM and all classical embeddings.}
\end{boxI}

\noindent\textbf{RQ-2:} \textit{Do LLM-based decontextualized embeddings capture similarity better than classical ones?}

%as the words being compared in BATS already have some inherent similarity over a random pair of words.

\smallskip

% To answer this question, we used the BATS dataset~\cite{gladkova-etal-2016-analogy}, which provides pairs of related words across many categories describing how those words are related. These categories were used to create two distinct sets of word pairs: 1) \textit{Morphologically Related Pairs} and 2) \textit{Semantically Related Pairs}. We also include a third set, \textit{Uncategorized Pairs}, which is made up of random word pairs from WordNet. 
{We evaluated word-pair similarity on the BATS dataset~\cite{gladkova-etal-2016-analogy}, categorizing pairs as \textit{Morphologically Related, Semantically Related}, or \textit{Uncategorized (random pairs)}. The uncategorized pairs are created using WordNet.} Figure~\ref{fig:sim-and-dissim-words} shows the distribution of cosine similarities for these categories across 11 embedding models.
% Next, we plot the distribution of cosine similarities between these pairs of words (by three categories) for all 11 embedding models (see Figure~\ref{fig:sim-and-dissim-words}).

Figure~\ref{fig:sim-and-dissim-words} shows that {Word2vec, GloVe, SBERT, and PaLM} exhibit the greatest separation between related pairs (both morphologically and semantically related) and unrelated pairs, which is the desired outcome. Other models, 
% struggle to differentiate between these categories
especially LLMs like OpenELM and GPT-ADA struggle to differentiate between categories, finding all more similar. 
% , which tend to find all categories more similar to each other. The remaining LLMs, except PaLM, also faced challenges in this differentiation. 
In contrast, classical models performed better at distinguishing morphological categories but did not perform well on semantic categories, as their distributions resembled those of random word pairs.

% In all models, morphologically related words were more similar to each other than semantically related ones (somewhat expected). USE, LLaMA, and LASER struggled to distinguish between semantically related pairs and random pairs.

\vspace{-2mm}
\begin{boxI}\small
\textit{\textbf{Finding-2}: LLMs are not always better than classical models in capturing semantic similarity (e.g., SBERT vs.~OpenELM). PaLM (LLM) and SBERT (Classical) can effectively distinguish semantically related and unrelated pairs, whereas most other models (both LLM-based and Classical) struggle with the same.}
\end{boxI}
\vspace{-3mm}

%to differentiate semantically related and unrelated pairs.

% \begin{table*}[!htb] \small
%     \centering
%     \begin{tabular}{rccccccc}
%     \toprule
%     Method & 3CosAdd & 3CosAvg & 3CosMul & LRCos & PairDistance & SimilarToAny & SimilarToB \\\midrule\midrule
%     SBERT & 0.243 & 0.261 & 0.267 & \textbf{0.487} & 0.086 & \color{blue}{\textbf{0.067}} & \textbf{0.141} \\
%     USE & 0.174 & 0.212 & 0.187 & 0.450 & 0.025 & 0.043 & 0.107 \\
%     LASER & 0.227 & 0.260 & 0.237 & 0.284 & 0.121 & 0.032 & 0.076 \\\hline
%     ADA-002 & \color{blue}{\textbf{0.412}} & \textbf{0.447} & \color{blue}{\textbf{0.424}} & 0.375 & \color{blue}{\textbf{0.232}} & 0.058 & \textbf{0.135} \\
%     LLaMA2 & 0.145 & 0.200 & 0.145 & 0.131 & 0.053 & 0.039 & 0.082 \\
%     PaLM 2 & \textbf{0.398} & \color{blue}{\textbf{0.458}} & \textbf{0.417} & \color{blue}{\textbf{0.534}} & \textbf{0.193} & \textbf{0.060} & 0.123 \\
%     \bottomrule
%     \end{tabular}
%     \caption{Results of BATS analogy task for each model by method.}
%     \label{tab:bert_analogy_performance_small}
% \end{table*}

\subsection{Analogy Task Based Comparison}
\textbf{RQ-3:} \textit{Do LLMs improve the performance of decontextualized word embeddings on analogy tasks?}
% \vspace{-3mm}

\smallskip
To answer this question, we followed the original word analogy task format set out by \citet{mikolov-etal-2013-linguistic} and comprehensively evaluated the eleven embedding models on the word pairs from the BATS dataset. For words $a,b,c,d$, analogy $a:b::c:d$ and embedding function $f(x)$, it is expected that $f(b) - f(a) + f(c) \approx f(d)$, which we will refer to as the \textbf{3CosAdd} method. Other approaches have been introduced for this task, including \textbf{Pair Distance} and \textbf{3CosMul} (introduced by \citet{levy-goldberg-2014-linguistic}). Later, \citet{drozd-etal-2016-word} introduced new methods called \textbf{3CosAvg} and \textbf{LRCos}, which achieved excellent performance in their experiments on classical models. For a detailed explanation, refer to appendix (Sec.~\ref{analogy_desc}).

\input{Tables/bats-analogy}
For all methods, the 3 words used as the input for the analogy were excluded from the answers, and top-1 accuracy was measured. For fairness, the same Wordnet corpus from section~\ref{similar-closeness} was used for each model, and the arithmetic results for each method were used to find the nearest neighbor in the corpus. These results are shown in Table \ref{tab:bert_analogy_performance_small}, with the best-performing embedding for each method shown in blue. Both ADA and PaLM performed very well, while OpenELM performed the worst in the LLM category. Among classical embeddings, SBERT and LASER performed quite well, often ranked higher than all open-source LLMs. Full information about each model's accuracy in each category can be found in the appendix (Table~\ref{tab:bats_analogy_performance}).

%LRCos seems to disproportionately benefit the classical models, leading to large accuracy increases for SBERT and USE, but this method also resulted in the highest overall accuracy when applied to PALM embeddings. The highest accuracy in each category across all models and methods is listed in bold. In general, models achieved better accuracy on the Morphology sections, with PALM and ADA seeming to do well with the non-baseline methods in general.

% \begin{boxI}\small
% \textit{\textbf{Finding-3}: 
% % Among LLMs, ADA and PALM perform significantly better than classical models on word analogy tasks, while SBERT \& LASER (classic models) are often ranked higher than all open-sourced LLMs, suggesting that smaller models can be an efficient alternative in resource-constrained devices.
% \end{boxI}
% \vspace{-4mm}

\vspace{-2mm}
\begin{boxI}\small
\textit{\textbf{Finding-3}: 
ADA and PALM outperform classical models on word analogy tasks. However, SBERT, GloVe, and Word2Vec often rank higher than open-source LLMs, indicating that smaller models can be efficient alternatives for resource-limited applications}
\end{boxI}
\vspace{-4mm}

% \begin{figure}[!htb]
%      \centering
%      \begin{subfigure}[b]{0.49\linewidth}
%          \centering
%          \includegraphics[width=\textwidth]{new_plots/Kendall.png}
%          \caption{Kendal $\tau$}
%          \vspace{-3mm}
%      \end{subfigure}
%      \hfill
%      \begin{subfigure}[b]{0.49\linewidth}
%          \centering
%          \includegraphics[width=\textwidth]{new_plots/Spearman.png}
%          \caption{Spearman $\rho$}
%          \vspace{-3mm}
%      \end{subfigure}
%     \vspace{-3mm}
%     \caption{Correlation coefficients for each pair of models, found using a large dataset of pairs of words.}
%     \label{fig:model-corr-coef}
%     \vspace{-6mm}
% \end{figure}

\subsection{Similarity Correlation Analysis}%Do LLMs Offer Something New?}

\textbf{RQ-4:} \textit{Do LLMs produce very different decontextualized word embeddings than the classical models?}

\begin{figure}[!htb]
    \vspace{-2mm}
    \centering
    \includegraphics[width=0.87\linewidth]{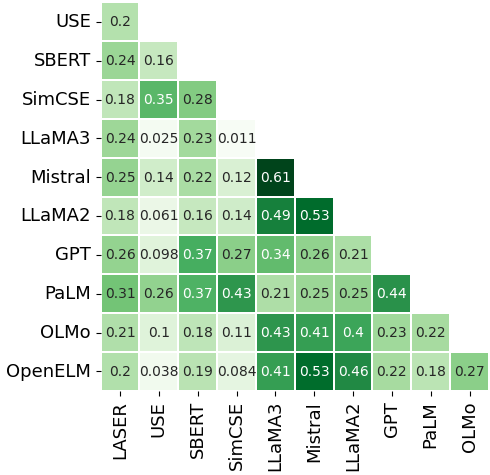}
    \vspace{-3mm}
    \caption{Spearman's $\rho$ for each model pair, calculated from $~2.1 B$ randomly selected word pairs out of a total of $6.4 B$ word pairs from the Wordnet (RQ1) corpus.}
    \label{fig:model-corr-coef}
    \vspace{-4mm}
\end{figure}

To further investigate whether LLMs offer something new/very different in terms of decontextualized embeddings, we computed statistical measures of correlation between each pair of models (both LLMs and Classical) in terms of their actual word embeddings. First, the cosine similarities of all pairs of words from the Wordnet corpus (see section~\ref{similar-closeness}) were computed for each embedding model. The correlation between two different embedding models was computed based on word pair similarities. Figure~\ref{fig:model-corr-coef} shows the Spearman's $\rho$ between each pair of embedding models (Kendall's $\tau$ correlation is reported in the appendix Figure~\ref{fig:model-corr-coef_apx} due to lack of space). Interestingly, these results show that the LLaMA family and Mistral are the most semantically similar, while SimCSE and LLaMA3 are the most different. Also, SimCSE and SBERT showed decent correlations with both ADA/PaLM. To ensure a fair comparison, Word2Vec and GloVe models were excluded due to their significantly different vocabulary sizes. %which would have hindered fair comparisons across all models.}

\begin{figure*}[!htb]
    \centering
    \includegraphics[width=0.9\linewidth]{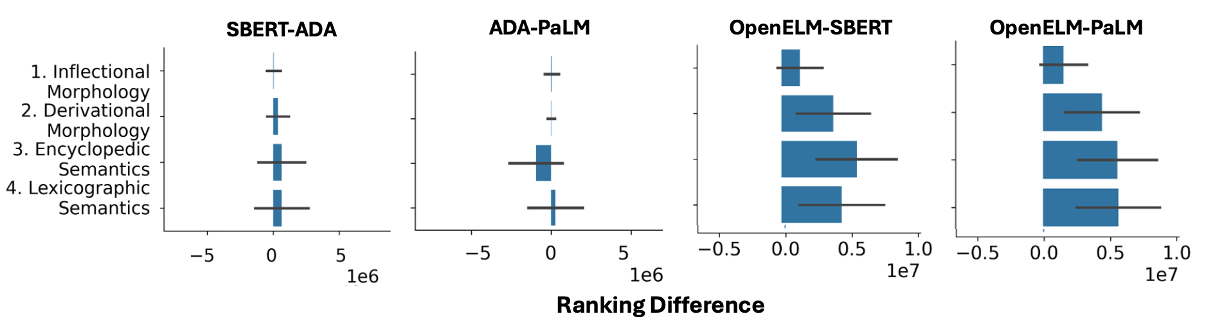}
    \vspace{-2mm}
    \caption{Mean-Variance plot of the difference in Word Pair Similarity Ranks for the BATS corpus. For all other model comparisons refer to appendix figure~\ref{fig:model-agreement-apx}.}
    \label{fig:model-rank-diff}
    \vspace{-4mm}
\end{figure*}

In another effort, we investigated how both types of models (LLMs and Classical) agreed/disagreed with each other regarding the similarity ranks of specifically related word pairs. More specifically, we computed the average difference of similarity ranks between pairs of words with three types of relations, morphological/semantic/random, for each pair of embedding models, where the rank is determined from the collection of all words in the BATS corpus (section~\ref{similar-closeness}). For example, if the 5th closest word to ``bad'' according to ADA-002's embedding was ``worst'', while ``worst'' was the 10th nearest word to ``bad'' according to LLaMA, we would compute a difference of $-5$ for that word pair while comparing ADA vs.~LLaMA. If two models mostly tend to agree on the similarity ranks of word pairs, we would expect an average value of $0$ with a small variance.

Figure~\ref{fig:model-rank-diff} presents these results for SBERT/ADA and ADA/PALM pairs (Figure~\ref{fig:model-agreement-apx} shows all pairs in the appendix due to lack of space), revealing that all models—except OpenELM—agree reasonably well on the similarity of words related by morphology. Notably, some model pairs such as PaLM-ADA, LLaMA3-LASER, and SBERT-ADA/PaLM exhibit greater agreement. It is surprising that ADA, PaLM, and SBERT demonstrate the highest levels of agreement despite substantial differences in model size and semantic space, suggesting that SBERT has a semantic space very similar to those of LLMs like ADA and PaLM.  In contrast, there were significantly more disagreements among the models for semantic relations.

\vspace{-2mm}
\begin{boxI}\small
\textit{\textbf{Finding-4}: Two LLMs, PaLM and ADA, tended to agree with each other in the decontextualized setting, additionally yielding a high correlation with SBERT, suggesting that SBERT is still an efficient choice when resources are constrained.}
\end{boxI}

% \begin{figure*}[!ht]
%     \centering
%     \includegraphics[width=\linewidth]{new_plots/context_decontextualized/polysemy_word.jpg}
%     \caption{Caption}
%     \label{fig:fig:poly_fig}
% \end{figure*}

 % All models except OpenELM tended to rank the related words higher than many models in nearly all categories. OpenELM and other models disagreed very strongly across all categories, suggesting that OpenELM embeddings are quite different from all others.

%In order to create a more direct comparison on the similarity of models in general, These similarities can now act as an annotated ``score'' for each word pair, and because the corpus of words is common, 
% \input{Tables/angle_table}
% \yash{since models have different dimensions, finding corr...}

%% file: Tables/bats-analogy.tex
\begin{table}[!hb]\LARGE
    \centering
    \begin{adjustbox}{width=0.95\linewidth}
    \begin{tabular}{rccccc}
    \toprule
\textbf{Method} & \textbf{3CosAdd} & \textbf{3CosAvg} & 
\textbf{3CosMul} & \textbf{LRCos} & \textbf{PairD} \\ \hline \hline
\textbf{GPT-ada}     & \color{blue}\textbf{0.4123} & \textbf{0.4465} & \color{blue}\textbf{0.4238} & 0.3750 & \color{blue}\textbf{0.2319} \\
\textbf{LLaMA2}   & 0.1449 & 0.2000 & 0.1454 & 0.1310 & 0.0526 \\
\textbf{LLaMA3}  & 0.0496 & 0.0590 & 0.0480 & 0.0530 & 0.0018 \\
\textbf{Mistral} & 0.0494 & 0.0620 & 0.0476 & 0.0635 & 0.0025 \\
\textbf{OLMo}    & 0.0525 & 0.0645 & 0.0499 & 0.0665 & 0.0018 \\
\textbf{OpenELM} & 0.0165 & 0.0350 & 0.0141 & 0.0135 & 0.0020 \\
\textbf{PaLM}    & \textbf{0.3981} & \color{blue}\textbf{0.4575} & \textbf{0.4171} & \color{blue}\textbf{0.5340} & \textbf{0.1929} \\ \hline
\textbf{SBERT}   & 0.2431 & 0.2605 & 0.2667 & \textbf{0.4870} & 0.0856 \\
\textbf{SimCSE}  & 0.0248 & 0.0385 & 0.0217 & 0.0315 & 0.0012 \\
\textbf{USE}     & 0.1739 & 0.2120 & 0.1873 & 0.4500 & 0.0251 \\
\textbf{LASER}   & 0.2271 & 0.2600 & 0.2369 & 0.2840 & 0.1214\\
\textbf{GloVe} & 0.3481 & 0.4290 & 0.3452 & 0.4875 & 0.1523\\
\textbf{Word2Vec} & 0.3229 & 0.3855 & 0.3096 & 0.4605 & 0.1203\\
\hline
% \bottomrule
\end{tabular}
\end{adjustbox}
\caption{Performance on BATS Analogy. \textbf{\color{blue}Blue} denotes the best accuracy; \textbf{black} denotes the second best. }
\label{tab:bert_analogy_performance_small}
\vspace{-3mm}
\end{table}

% \begin{table}[!htb]
%     \vspace{-4mm}
%     \centering
%     \begin{adjustbox}{width=\linewidth}
%     \begin{tabular}{rccccc}
%     \toprule
%     Method & 3CosAdd & 3CosAvg & 3CosMul & LRCos & PairD \\\midrule\midrule
%     SBERT & 0.243 & 0.261 & 0.267 & \textbf{0.487} & 0.086 \\
%     USE & 0.174 & 0.212 & 0.187 & 0.450 & 0.025\\
%     LASER & 0.227 & 0.260 & 0.237 & 0.284 & 0.121 \\\hline
%     ADA-002 & \color{blue}{\textbf{0.412}} & \textbf{0.447} & \color{blue}{\textbf{0.424}} & 0.375 & \color{blue}{\textbf{0.232}}\\
%     LLaMA2 & 0.145 & 0.200 & 0.145 & 0.131 & 0.053\\
%     PaLM 2 & \textbf{0.398} & \color{blue}{\textbf{0.458}} & \textbf{0.417} & \color{blue}{\textbf{0.534}} & \textbf{0.193}\\
%     \bottomrule
%     \end{tabular}
%     \end{adjustbox}
%     \caption{Performance on BATS Analogy. Blue denotes the best accuracy; bold black denotes the second best.}
%     \vspace{-4mm}
%     
% \end{table}

%% file: 4.1_Contextualized.tex
\section{Comparing Contextual Embeddings: LLM vs.~Classical}
\vspace{-1.5mm}
In the contextualized setting, we compare LLM vs.~Classical word/sentence embeddings across nine different variational tasks. This way allows us to examine how context influences different embedding models across various linguistic scenarios\footnote{Contextualized embeddings are generated by processing sentences from the nine contextual tasks and extracting the embeddings corresponding to the anchor words. Due to API limitations, closed-source models like GPT-ADA-002 and PaLM2-Gecko were excluded from the contextualized analysis. Similarly, classical models such as USE and LASER, which do not readily provide contextualized word embeddings, were omitted from this part of the study.}. The variational tasks include:

\vspace{-2mm}
\subsection{Variational Tasks}\label{task}
\begin{itemize}[leftmargin=*,itemsep=0ex,partopsep=-1ex,parsep=0ex]
\item \textbf{Lexical Variations:}

\begin{itemize}[leftmargin=*,itemsep=0ex,partopsep=-1ex,parsep=0ex]
    
    \item \textbf{Synonym Task}: Generate sentence $S_1$ containing an anchor word. Create $S_2$ by replacing a word before the anchor word in $S_1$  with its \textit{synonym}. Compare the anchor word embeddings from both sentences.
    
    \item  \textbf{Antonym Task}: Similar to the Synonym Task, but replace the word with its \textit{antonym}.

    \item  \textbf{Negation Task}: Generate $ S_2$ by adding a \textit{negation} before the anchor word in $S_1$. Compare the anchor word embeddings.
\end{itemize}

\item \textbf{Tone Variations:} First, Generate $S_1 $ with an anchor word, and then -

\begin{itemize}[leftmargin=*,itemsep=0ex,partopsep=-1ex,parsep=0ex]
    
    \item  \textbf{Exclamation Task}: Create four \textit{exclamatory} variations of $S_1 $ with the anchor word. Compare the anchor word embeddings.

    \item  \textbf{Question Formation Task}: Create four \textit{interrogative} sentences based on $S_1 $ containing the anchor word. Compare the anchor word embeddings.

    \item  \textbf{Active-Passive Task}: Generate $S_1$ in \textit{active voice}. Create four \textit{passive voice} versions of $S_1 $, keeping the anchor word. Compare the anchor word embeddings.
    
    \end{itemize}
    
\item \textbf{Semantic Variations:}

\begin{itemize}[leftmargin=*,itemsep=0ex,partopsep=-1ex,parsep=0ex]
    \item \textbf{Jumbling Task}: Generate $S_1 $ with an anchor word. Create the following sentences by:\\
        - $S_2 $: Shuffling words before the anchor word.\\
        - $S_3 $: Shuffling the entire sentence.\\
        - $S_4 $ and $S_5 $: Exchanging one or two words around the anchor word.\\
    Finally, compare the anchor word embeddings. %across these sentences.

    \item  \textbf{Paraphrasing Task}: Generate $S_1 $ with an anchor word. Create four \textit{paraphrases} of $S_1 $, all containing the anchor word. Compare the anchor word embeddings across these sentences.
    
    \item \textbf{Polysemy Task}: Generate five sentences using the anchor word in different \textit{senses} (polysemy). Compare the embeddings to assess how models capture multiple meanings.
\end{itemize}
\vspace{-0.5mm}
\end{itemize}
{Due to LLMs' causal attention mechanism, we applied all variations before the anchor word, except for jumbling. Since causal attention computes embeddings based on preceding words, this ensures the perturbations influence the anchor word's embedding.}
Next, for each variational task, we compute 3 different similarity scores, as follows.

\textbf{1) Anchor Inter-Contextual Variance}: Here, we measure the variance of anchor word embeddings across different contexts. First, we extracted the embedding of each anchor word from all generated sentences. We then designated the embedding from the first sentence as the reference embedding. Subsequently, we computed the cosine angle between this reference embedding and the anchor word embeddings from the remaining sentences. The average of these cosine angles quantifies how differently the model represents the anchor word across various contexts.

% To evaluate word embeddings in the contextualized setting, we employed a comparative approach called \textit{inter-contextual similarity}. For each anchor word, we extracted its embedding from all generated sentences. We then designated the embedding from the first sentence as the reference embedding. Subsequently, we computed the inter-contextual cosine similarity between this reference embedding and the anchor word embeddings from the remaining sentences. The average of these similarity scores served as the final similarity score for each anchor word. This inter-contextual analysis allows us to quantify how consistently the models represent the anchor word across various contexts. 

% with higher scores indicating greater consistency in representation despite contextual changes.

% how we compared decontextualized vs contextualized word embedding

\textbf{2) Anchor Contextual Deviation}: Here, we compute the cosine angle between the standalone (decontextualized) anchor word embedding and the anchor word contextual embeddings extracted from each generated sentence. We then averaged these cosine angles to obtain a measure of how much the contextualized representations deviate from the decontextualized ones.

\textbf{3) Sentence Meaning Variance}: Here, we measure how the sentences overall are semantically similar/different by computing the cosine angle between them. The average cosine angle between two sentence embeddings is reported.

% To maintain data quality, we apply a filtering process, excluding anchor words if (1) the generated sentences do not contain the anchor word, or (2) the anchor word is altered during sentence generation. This ensures the integrity of our analysis.

% We generated both decontextualized and contextualized embeddings from each model. Decontextualized embeddings were created by inputting single words into each model's tokenizer. For models using single-token inputs, we utilized the final hidden state. In cases of sub-word tokenization, we averaged the final hidden states of each token. Contextualized embeddings were generated by inputting sentences from our proposed nine contextual tasks. We extracted in-context anchor word embeddings by locating the position of anchor words in the sentences and retrieving their respective embeddings from the final hidden states.

% Due to API limitations, closed-source models like GPT-ADA-002 and PaLM-Gecko were excluded from the contextualized embedding analysis. Similarly, classical models such as USE and LASER, which do not generate contextualized word embeddings, were omitted from this part of the study.
% \smallskip
\vspace{-2mm}
\subsection{Dataset Generation}
\vspace{-1.5mm}
To facilitate our contextual analyses, we created a synthetic dataset by randomly sampling $1,200$ anchor words (nouns, verbs, or adjectives) from WordNet. We then used the Claude-sonnet 3.5 model~\cite{claude} to generate sentences for each variational task based on these words, ensuring a diverse and comprehensive set of contextual scenarios. The prompts to generate the dataset are shown in the appendix section~\ref{prompting}.

For lexical variational tasks,  we generated only two sentences (one reference and one variational) for each anchor word, as have a very high word overlap between sentences. For the remaining six categories, we created five sentences for each anchor word (refer to Section~\ref{task}). Each set of sentences shared the same anchor word, but in different contexts (see examples in Appendix~\ref{tab:samples}). %This approach allows us to examine the models' behavior across various complex contextual scenarios.

To compute cosine angles, we extracted three types of embeddings: 1) Decontextualized anchor word embeddings from each model. 2) Contextualized anchor word embeddings (token-level anchor word embedding from the last hidden layer of each model), and 3) Sentence embeddings (overall embedding for each generated sentence). This multi-faceted approach allows us to compare word representations in both contextualized and decontextualized settings across different models and variational tasks, providing a nuanced understanding of each model's strengths and limitations.

%% file: 5results.tex
\input{Tables/angle_table_2}
\vspace{-2mm}
\subsection{Research Questions and Findings}

\noindent\textbf{RQ-5:} \textit{How do LLMs differ from classical embeddings for single lexicon variations?}

To examine how models handle single lexicon variations we analyze the \textit{Synonym, Antonym, and Negation} variational tasks and compare cosine angle (see Table~\ref{tab:cos_angle}). These tasks modify sentences by replacing a word with its synonym or antonym or by introducing a negation before the anchor word, which affects contextual understanding. 
% For comparison, we compute the cosine angles on all three fronts (see Table~\ref{tab:cos_angle}).

% between anchor word embeddings and sentence embeddings was calculated.
%shorten the paragraph
{For all variations (\textit{Synonym, Antonym, and Negation}), we expect a high value for \textit{Anchor Contextual Deviation} (i.e., contextual word embeddings should be somewhat different from the corresponding decontextualized ones), and found LLaMA2 excelling in this aspect.} 
% Interestingly, particularly LLaMA2 excelled in differentiating between context-aware and context-free embeddings of anchor words (as shown by the highest Anchor Contextual Deviation in Table~\ref{tab:cos_angle}).

For synonym variations, we expect a low value for \textit{Anchor Inter-contextual Variance} and \textit{Sentence Meaning Variance}, as the overall meanings are typically unaltered. Our experimental results align with these expectations, with the classical model SimCSE showing the lowest cosine angle (low variance) in the \textit{Inter-Contextual} setting. For antonyms and negations, we anticipate larger variance (cosine angles) due to flipped/opposite meanings, which is not quite manifested in Table~\ref{tab:cos_angle} for any model. Perhaps high word overlap between sentence pairs caused models to overlook a single lexicon difference, resulting in lower-than-expected variance (cosine angles). This observation aligns with findings from~\cite{mahajan-etal-2024-align}, where the authors demonstrated that high word overlap could lead models to ignore subtle differences.

Also, when comparing antonym tasks to synonym tasks, we observed increased angles across all models, indicating some sensitivity to opposite meanings. SimCSE showed the highest percentage change in angle ($\sim50\%$), demonstrating strong antonym differentiation, while OpenELM exhibited a smaller change ($\sim 15\%$), suggesting potential struggles with antonym variations.
For the negation task, the addition of negation words resulted in relatively higher angles, indicating their ability to capture the influence of negation on sentence meaning to some extent in varying degrees.

%At the sentence level, all models effectively captured similarities between synonymous sentences. 

\vspace{-2mm}
\begin{boxI} \small
    \textit{\textbf{Finding-5:} LLaMA models, particularly LLaMA2, excelled in Anchor Contextual Deviation for single lexicon variations. OLMo achieved superior Anchor Inter-Contextual Variance in Antonym and Negation tasks, while SBERT achieved superior Sentence Meaning Variance for the same.}
\end{boxI}

\noindent\textbf{RQ-6:} \textit{How do LLMs differ from classical embeddings for linguistic tone variations?}

We examine the \textit{Exclamatory, Questionnaire, and Active-Passive} variational tasks, each involving five sentences per anchor word. The first sentence is the reference generated using the anchor word, while the remaining four are tailored to each category, sharing the anchor word in common. For these tasks, wider angles are desired for \textit{Anchor Contextual Deviation}, but, lower angles for \textit{Anchor Inter-Contextual Variance} and \textit{Sentence Meaning Variance} (similar to the synonym task) as these are just tonal variations of the reference.

% Note: How to write: OpenELM is not a good model as it always computes high cos sim. Table 2 showcases that openelm is best in semantically similar meaning categories, how that is flawed as it also gives low scores in polysemy, negation, and antonym. If we remove the openelm model then we can say that the classical model SIMCSE is the best model on sentence level. 

%We anticipate that contextualized anchor words will yield similar embeddings due to their semantic similarity, while contextual deviated words will show dissimilarities.

Table~\ref{tab:cos_angle} demonstrates that LLaMA2 again excels in \textit{Anchor Contextual Deviation} (widest angles), while classical models (SBERT and SIMCSE) yield low cosine angles in \textit{Anchor Inter-Contextual Variance}. Interestingly, OpenELM achieved low \textit{Sentence Meaning Variance} for \textit{Questions} and \textit{Exclamations}.  Notably, all LLMs (except OpenELM) exhibit slightly higher \textit{Sentence Meaning Variance} than the SimCSE model. The corresponding cosine similarity spread is illustrated in Figures~\ref{active_passive_plot},\ref{para_plots},\ref{exclamatory_plots},\ref{question_plots} in the appendix (due to lack of space), which further confirms these findings.

\vspace{-2mm}
\begin{boxI}\small \textit{\textbf{Finding 6:} In the context of tone variations, SimCSE and SBERT achieve lower Inter-Contextual Variance for anchor words (a desired behavior), while openELM among LLMs exhibits lower Sentence Meaning Variance (another desired behavior). Finally, LLaMA models, particularly LLaMA2, excelled in \textit{Anchor Contextual Deviation} in the context tone variations, yielding wider angles.} \end{boxI}

\noindent\textbf{RQ-7:} \textit{How do LLMs differ from classical embeddings for overall semantic variations?}

We computed the cosine angles across all three fronts (\textit{Inter-Contextual Variance}, \textit{Anchor Contextual Deviation}, and \textit{Sentence Meaning Variance}) for the 3 variational tasks: \textit{Jumbling}, \textit{Paraphrasing} and \textit{Polysemy}. In all these tasks, wider angles are desired for all 3 measures across all 3 tasks, with the only expectation that lower angles are desired for \textit{Anchor Inter-Contextual Variance} and \textit{Sentence Meaning Variance}  in the case of \textit{Paraphrasing} task. %(similar to the synonym task).

%capturing subtle context-induced meaning shifts. 

Consistent with previous findings, LLaMA2 achieved the highest Anchor Contextual Deviation for all tasks, as seen in Table~\ref{tab:cos_angle}. In fact, LLaMA2 performed the best across all three variance measures for the Jumbling task, suggesting its superior capability in capturing word order. All models demonstrate somewhat high Inter-Contextual Variance for polysemous word context (a desired behavior), with OLMo performing particularly well, suggesting it is adept at detecting polysemy. Finally, results were mixed for the paraphrasing task. %without a clear winner. across all three measures. 

%This contrast is likely due to differences in architecture and training objectives: LLMs, with larger embeddings and next-word prediction training, excel in word-level context capture, while classical encoder-only models, optimized for sentence encoding, are better at capturing sentence-level context.

%On the , the LLaMA2 dominates by showcasing a wider cosine angle between word vectors in different contexts. 

%Interestingly, while classical models have comparative performance with token-level context for polysemous words, they outperform LLMs in sentence-level comparisons despite having fewer parameters. 

% Models with high decontextualized similarity scores may still struggle with context-sensitive variations. The performance gap between token- and sentence-level tasks suggests that different architectural features are at play. Classical models' success in sentence-level tasks indicates that sentence aggregation can mask polysemy challenges, while LLMs excel in token-level disambiguation. These results underscore the challenge of building models that effectively capture both fine-grained and general contexts.

% Overall, while models generally perform as expected in these tasks, classical models struggle more with token-level nuances but excel at sentence-level comparisons. LLMs, particularly LLaMA2, demonstrate superior performance in capturing fine-grained contextual differences at the token level.

\begin{boxI} \small \textit{\textbf{Finding-7}: LLaMA2 performed the best in capturing word orders (i.e., Jumbling task). In the case of Polysemy, classical models outperform LLMs in sentence-level similarity, while LLMs (especially OLMo) excel at token-level disambiguation, highlighting a trade-off between fine-grained and overall context understanding. Finally, paraphrasing task results were mixed.} \end{boxI}

%% file: Tables/angle_table_2.tex
\begin{table*}[!htb]
\centering
\small
\begin{adjustbox}{width=0.78\textwidth}
\begin{tabular}{l|ccc|ccc|ccc}
\toprule
\textbf{Lexical} & \multicolumn{3}{c|}{\textbf{Synonym}} & \multicolumn{3}{c|}{\textbf{Antonym}} & \multicolumn{3}{c}{\textbf{Negation}} \\
\cmidrule{2-10}
\textbf{Variations} & \textbf{Inter.} $\downarrow$& \textbf{Deviation }$\uparrow$& \textbf{Sim.} $\downarrow$
& \textbf{Inter.} $\uparrow$& \textbf{Deviation }$\uparrow$& \textbf{Sim.} $\uparrow$
& \textbf{Inter.} $\uparrow$& \textbf{Deviation }$\uparrow$& \textbf{Sim.} $\uparrow$\\
\midrule
\textbf{SBert} & 10.74 & 45.69 & 18.13 & 18.45 & 46.64 & \color{blue}\textbf{27.21} & 24.41 & 47.53 & \color{blue}\textbf{38.48}  \\
\textbf{SimCSE} & \color{blue}\textbf{9.87}  & 47.77 & \textbf{9.39} & 22.33 & 49.07 & 21.00 & 29.12 & 50.92 & \textbf{26.75} \\
\cmidrule{1-10}
\textbf{LLaMA3} & 15.26 & \textbf{69.41} & 12.40 & 22.89 & 67.79 & 17.04 & \textbf{30.42} & \textbf{69.74} & 21.84 \\
\textbf{LLaMA2} & 15.21 & \color{blue}\textbf{81.99}  & 12.94 & 22.23 & \color{blue}\textbf{78.15}  & 16.76 & 28.93 & \color{blue}\textbf{80.58}  & 22.80 \\
\textbf{Mistral} & 15.14 & 60.13 & 11.15 & \textbf{22.95} & 59.40 & 14.87 & 28.77 & 59.55 & 19.70 \\
\textbf{OLMo} & 16.51 & 58.62 & 13.78 & \color{blue}\textbf{25.10}  & 57.37 & 18.66 & \color{blue}\textbf{31.51}  & 57.26 & 24.50 \\
\textbf{OpenELM} & \textbf{10.33} & 68.23 & \color{blue}\textbf{8.58}  & 16.89 & \textbf{67.90} & {9.88} & 20.18 & 68.41 & {13.01} \\
\bottomrule

\multicolumn{10}{c}{} \\

\multicolumn{10}{c}{
\begin{tabular}{l|ccc|ccc|ccc}
\toprule
\textbf{Tone} & \multicolumn{3}{c|}{\textbf{Exclamatory}} & \multicolumn{3}{c|}{\textbf{Questionnaire}} & \multicolumn{3}{c}{\textbf{Active-Passive}} \\
\cmidrule{2-10}
\textbf{Variations} & \textbf{Inter. }$\downarrow$& \textbf{Deviation }$\uparrow$& \textbf{Sim.} $\downarrow$
& \textbf{Inter.} $\downarrow$& \textbf{Deviation }$\uparrow$& \textbf{Sim.} $\downarrow$
& \textbf{Inter.} $\downarrow$& \textbf{Deviation }$\uparrow$& \textbf{Sim.} $\downarrow$\\
\midrule
\textbf{SBert} & \color{blue}\textbf{23.81} & 44.89 & 38.61 & \color{blue}\textbf{21.28} & 45.05 & 33.01  & \textbf{20.24} & 44.76 & 25.12\\
\textbf{SimCSE} & \textbf{24.52} & 47.64 & \textbf{27.00} & \textbf{21.75} & 47.19 & \textbf{21.74} & \color{blue}\textbf{18.13} & 47.53 & \color{blue}\textbf{15.51}\\
\cmidrule{1-10}
\textbf{LLaMA3} & 38.66 & 64.76 & 30.34 & 39.53 & 63.22 & 30.45 & 43.65 & 65.80 & 27.09\\
\textbf{LLaMA2} & 39.60 & \color{blue}\textbf{71.21}  & 30.78 & 38.87 & \color{blue}\textbf{69.50} & 29.46 & 45.82 & \color{blue}\textbf{73.90}  & 27.94\\
\textbf{Mistral} & 35.80 & 55.68 & 27.71 & 36.65 & 56.21 & 26.74 & 41.00 & 57.29 & 24.01\\
\textbf{OLMo} & 42.85 & 54.74 & 34.54 & 44.01 & 54.96 & 33.06 & 46.85 & 56.60 & 31.10 \\
\textbf{OpenELM} & 27.54 & \textbf{67.04} & \color{blue}\textbf{19.50}  & 27.99 & \textbf{67.10} & \color{blue}\textbf{17.39}  & 29.65 & \textbf{66.74} & \textbf{15.53} \\
\bottomrule
\end{tabular}
}\\
\multicolumn{10}{c}{} \\

\multicolumn{10}{c}{
\begin{tabular}{l|ccc|ccc|ccc}
\toprule
\textbf{Semantic} & \multicolumn{3}{c|}{\textbf{Polysemy}} & \multicolumn{3}{c|}{\textbf{Paraphrase}} & \multicolumn{3}{c}{\textbf{Jumbling}} \\
\cmidrule{2-10}
\textbf{Variations} & \textbf{Inter. }$\uparrow$& \textbf{Deviation }$\uparrow$& \textbf{Sim.}$\uparrow$
& \textbf{Inter. }$\downarrow$& \textbf{Deviation }$\uparrow$& \textbf{Sim.} $\downarrow$
& \textbf{Inter. }$\uparrow$& \textbf{Deviation }$\uparrow$& \textbf{Sim.} $\uparrow$\\
\midrule
\textbf{SBert} & 46.33 & 56.38 & \color{blue}\textbf{75.49}  & \textbf{26.05} & 45.11 & 42.59 & 17.41 & 51.40 & 19.45 \\
\textbf{SimCSE} & 54.62 & 58.59 & \textbf{57.81} & \color{blue}\textbf{24.99}  & 47.98 & \textbf{26.16} & 17.56 & 51.63 & 15.03  \\
\cmidrule{1-10}
\textbf{LLaMA3} & 55.64 & \textbf{78.97} & 52.49 & 39.22 & 65.20 & 27.17 & \textbf{52.86} & \textbf{73.19} & \textbf{38.32} \\
\textbf{LLaMA2} & \textbf{59.51} & \color{blue}\textbf{88.18}  & 48.16 & 40.61 & \color{blue}\textbf{73.40}  & 26.44   & \color{blue}\textbf{56.75} & \color{blue}\textbf{73.38} & \color{blue}\textbf{51.89} \\
\textbf{Mistral} & 58.60 & 71.09 & 41.95 & 37.48 & 57.18 & 25.88 & 42.68 & 63.39 & 27.91 \\
\textbf{OLMo} & \color{blue}\textbf{63.58}  & 67.90 & 55.07 & 43.18 & 55.83 & 31.16 & 47.95 & 60.04 & 34.68 \\
\textbf{OpenELM} & 51.13 & 71.12 & 28.77 & 28.41 & \textbf{67.13} & \color{blue}\textbf{20.05}   & 29.65 & 66.74 & 15.53\\
\bottomrule
\end{tabular}
} 
\end{tabular}
\end{adjustbox}
\caption{Comparison of different models across various tasks in the Contextualized Evaluation setting. The values represented are the \textbf{Average Cosine Angle}. Arrows ($\uparrow \downarrow$) indicate expected behavior: $\uparrow$ suggests a lower cosine angle is desirable, and $\downarrow$ is the opposite.  The lower the angle, the higher the cosine similarity.  Here, \textbf{`Inter.'} represents \textbf{Anchor Inter-Contextual Variance}, 
% measuring the average cosine similarity between the standalone anchor word embedding and its contextualized embeddings. 
\textbf{`Deviation'} represents \textbf{Anchor Contextual Deviation}, 
% indicating how much the contextualized embeddings deviate from the decontextualized one. 
\textbf{`Sim'} stands for \textbf{Sentence Meaning Variance}, 
% measuring the similarity of sentence embeddings across different contexts. 
The \textbf{\color{blue}best} and \textbf{2\textsuperscript{nd}} best scores in each category are highlighted in respective colors.}
\label{tab:cos_angle}
\vspace{-3mm}
\end{table*}

%% file: 6conclusion.tex
\vspace{-3mm}
\section{Discussions and Final Words}
\vspace{-2mm}
In this paper, we compared word/sentence embeddings from 7 LLMs and 6 classical models (total 13) in both contextualized and decontextualized settings. %We evaluated a total of 13 models, including 7 LLMs and 6 classical models. 
In the decontextualized setting, we used WordNet and the BATS dataset to create a corpus of 80,000 unique words and 6.4 billion word pairs. 
% Notably, static embedding models like Word2Vec and GloVe were excluded from some analyses due to their limited vocabulary sizes.
% \yash{It's important to note that static embedding models like Word2Vec and GloVe were excluded from some analyses due to their limited vocabulary sizes.} 
Our results show that LLM-based models PaLM and ADA performed the best on word analogy tasks, surprisingly aligning with SBERT, suggesting SBERT as a resource-efficient alternative.  
% . This suggests that while PaLM and ADA are indeed powerful, SBERT offers a resource-efficient alternative without significant loss in accuracy.
% and demonstrated a surprising alignment with SBERT. This suggests that while PaLM and ADA are indeed powerful, SBERT offers a resource-efficient alternative without significant loss in accuracy.

% We designed nine evaluation tasks using synthetic datasets derived from nearly $1,200$ anchor words. For each anchor word, we generated multiple sentences using the Claude-Sonnet model and calculated cosine similarity across all tasks. Our findings reveal that while all models capture contextualized word embeddings to some degree, LLMs excel in contextual deviation settings, demonstrating their strength in token-level analysis. Conversely, classical models like SimCSE outperform LLMs in sentence-level tasks, underscoring their ongoing relevance for fine-grained semantic analysis.
%We designed nine linguistic tasks using synthetic datasets derived from approximately 1,200 anchor words. For each anchor word, we generated multiple sentences using the Claude-Sonnet model and calculated cosine similarities across all tasks on three fronts: \textit{Anchor Inter-Contextual Variance, Anchor Contextual Deviation, and Sentence Meaning Variance}.

In the contextualized setting, we assessed 5 LLMs and 2 classical models 
% (since some models did not produce contextualized embeddings or were closed source) 
across three variance measures: \textit{Anchor Inter-Contextual Variance, Anchor Contextual Deviation, and Sentence Meaning Variance} across 9 variational tasks. We found LLMs (especially LLaMA2) excel in \textit{Anchor Contextual Deviation} across all contexts, demonstrating superior contextualized token-level analysis. Conversely, classical models (SimCSE and SBERT) outperformed many LLMs in terms of \textit{Sentence Meaning Variance} for lexicon variation and \textit{Polysemy tasks}, underscoring their continued relevance. Interestingly, OLMo achieved superior \textit{Anchor Inter-Contextual Variance} in Antonym, Negation, and Polysemy tasks, demonstrating its superiority in properly contextualizing word embeddings in flipped-meaning scenarios.

%a trade-off between capturing fine-grained context and maintaining overall semantic consistency. In particular, we found that while LLMs, especially LLaMA2,
% all models capture Inter-contextual variance to some extent, LLMs (especially LLaMA2) 

\vspace{-2mm}
\subsection{Why LLMs Differ from Classics?}
While a detailed root-cause analysis of our findings is beyond the scope of this paper, we present two hypotheses that may explain the performance disparities between models. 

\begin{itemize}[leftmargin=*,itemsep=0.2ex,partopsep=-1ex,parsep=0ex]
%excel at generalization and semantic reasoning but often

\item \textbf{First}, LLMs like GPT-3 and LLaMA, due to being trained on massive and diverse datasets may often over-generalize, resulting in high similarity scores for unrelated word pairs. 

%than for fine-grained relational tasks. In contrast

\item \textbf{Second}, While, LLM's autoregressive or masked language modeling objectives are optimized for text generation, classical models like SBERT and USE, are trained on task-specific datasets with contrastive loss, which may make them particularly effective for contrastive tasks like a synonym or antonym replacement. 

%\item \textbf{Lastly}, from an architectural perspective, encoder-based models like SBERT are specifically designed for sentence embeddings, enabling greater precision in semantic similarity tasks, while decoder-based LLMs focus on generative capabilities.

\end{itemize}

%As word/sentence embeddings represent a word/sentence through a real-valued vector in a high-dimensional space, it is natural for humans to expect these vectors to follow particular vector similarity properties under a specific semantic transformation. 

%In other words, humans would expect a higher cosine angle between the contextual and decontextualized embedding.

\vspace{-2mm}
\subsection{Implications and Future Use}

\begin{itemize}[leftmargin=*,itemsep=0.2ex,partopsep=-1ex,parsep=0ex]
\item \textbf{Interpretability of Embeddings}: Consider the Antonym replacement task. It is natural for humans to expect the contextual word embedding of the anchor word to be somewhat distinct from the corresponding decontextualized ones, as antonyms significantly change the semantics of the sentence.  If a particular embedding model indeed adheres to this expectation, it would be ``easy to interpret'' for humans. In this way, our analysis can help quantify and compare the ``interpretability'' of various embedding models.

%and, hence, can enable quantitative comparison across various models in terms of their interpretability.

%Encoding models are expected (by humans) to yield similar embeddings for word/sentence pairs with high semantic overlap and distinct embeddings for word/sentence pairs with substantial semantic deviation. At least, that is how humans learn and interpret the similarity of real-valued vectors in coordinate geometry. Our analysis tests whether existing embedding models align with human expectations of similarity. 

\item \textbf{Accuracy Vs.~Interpretability:}  Our findings reveal that a model's interpretability (i.e., alignment with human expectation) varies widely depending on the specific contextual transformation and degree of semantic deviation. In other words, we have yet to discover the ``ideal'' encoder that can guarantee both accuracy and interpretability.

%at the same time, aligns well with human expectations.

\item \textbf{Embedding Use in Specific Scenarios:} Our comparative analysis can help practitioners choose between LLM-induced and classical embeddings for specific scenarios. For example, LLMs (e.g., ADA) excel in clustering semantically related words in decontextualized settings, while classical models like SimCSE yield more accurate embeddings for contextualized sentence-level similarity tasks. %As such, the practitioner/engineer can choose a particular model depending on the task and context.
\end{itemize}

%% file: 7Limitation.tex
\section{Limitations}

Despite providing a comprehensive comparison between classical embedding models and Large Language Models (LLMs) in both decontextualized and contextualized settings, our study has several limitations. First, due to computational constraints and API restrictions, we were unable to include some closed-source models and larger LLMs in the contextualized embedding analysis, which may limit the generalizability of our findings across all state-of-the-art models. Second, our evaluation focuses solely on the English language and uses synthetic sentences generated by the Claude-Sonnet model, which may not capture the full diversity and complexity of natural language in real-world contexts. Third, while we explored a range of linguistic tasks, this represents only a subset of the wide spectrum of linguistic evaluations that could be incorporated into future extensions of this framework.

Moreover, numerous works~\cite{linzen2016issues,fournier2020analogies} have highlighted issues with using the standard analogy task to determine if semantic information is encoded in word embeddings. Therefore, we have refrained from making claims that one embedding is inherently "better" than another. Additionally, our reliance on cosine similarity as the primary metric assumes it adequately reflects semantic similarity between embedding vectors. While it is a popular choice in NLP literature, cosine similarity has inherent limitations, and our findings are constrained by this methodological assumption.

% This work only analyzes six models due to time and resource constraints. More work will need to be done to increase the number of models tested to draw a more general conclusion on the difference between Classical and LLM-based embeddings. Additionally, this work limits its investigations to the English language.

% Numerous works~\cite{linzen2016issues,fournier2020analogies} have illuminated issues in using the standard analogy task to determine if semantic information is encoded in word embeddings. Therefore, this paper has refrained from making statements implying that one embedding is `better' than another. 

% This work also relies upon the assumption that cosine similarity is a reasonable metric to compute semantic similarity between two word embedding vectors. While the interpretation of cosine similarity can itself be debatable, cosine similarity is still the most popular similarity metric used in NLP literature. However, our findings are limited by the inherent limitations of the cosine similarity metric.

% add points:
\begin{comment}
    - when fetching anchor word embeddings it does consider further words for context unitl last layers. 
    - 
\end{comment}

%% file: 8appendix.tex
\appendix
\section{Appendix}
\section{Decontextualized Evaluation Setting}
\subsection{Analogy-Task Based Comparison}\label{analogy_desc}
Here we presented the exhaustive list of model accuracy on various evaluation methods of the Analogy task. 
See Table \ref{tab:bats_analogy_performance} for a more granular description of the performance of each model on specific categories of BATS. Here is the description of the Metric we used to evaluate the analogy task:
% \begin{itemize}
%     \item \textbf{3CosAdd}
%     \item  \textbf{3CosAvg}
%     \item \textbf{3CosMul}
%     \item \textbf{LRCos}
%     \item \textbf{PairDistance}
%     \item \textbf{SimilarToAny}
%     \item \textbf{SimilarToB}
    
% \end{itemize}
\begin{enumerate}
    \item \textbf{3CosAdd}: \\
    The analogy $a:b::c:d$ is solved by computing $f(b) - f(a) + f(c) \approx f(d)$. For example, in the analogy "king:man::queen:woman", the equation becomes $f(\text{man}) - f(\text{king}) + f(\text{queen}) \approx f(\text{woman})$.

    \item \textbf{3CosAvg}: \\
    This extends 3CosAdd by averaging the transformations over multiple analogy pairs. For "king:man::queen:woman", we take the average of multiple such pairs to improve accuracy:
    \[
    f(d) \approx \text{avg}(f(b) - f(a) + f(c)).
    \]

    \item \textbf{3CosMul}: \\
    Similar to 3CosAdd but instead of adding, it multiplies cosine similarities:
    \[
    \text{argmax}_d \ \frac{\text{cos}(f(b), f(d)) \cdot \text{cos}(f(c), f(d))}{\text{cos}(f(a), f(d)) + \epsilon}.
    \]

    \item \textbf{LRCos}: \\
    A method using logistic regression to classify whether the analogy holds, using distances between embeddings.

    \item \textbf{PairDistance}: \\
    Measures the cosine distance between two pairs of words $(a, b)$ and $(c, d)$ to check how similar their relationship is. For "king:queen", the cosine distance is compared with "man:woman".

    \item \textbf{SimilarToAny}: \\
    Checks if $d$ is similar to any of the words in the analogy $(a, b, c)$. For "king:man::queen:?", it checks if $f(d)$ is similar to any of "king", "man", or "queen".

    \item \textbf{SimilarToB}: \\
    Checks if $d$ is most similar to $b$ in the analogy. For "king:man::queen:?", the method finds the word most similar to "man".
\end{enumerate}

Below Table~\ref{fig:model-agreement-apx} showcase the extensive comparison of all the models on analogy task using various evaluation metrics.

The following sections in the appendix are organized as follows: Section~\ref{analogy_ranking_apx} presents the ranking comparison of models on the Word Analogy Task. Section~\ref{sim_corr} provides Kendall's $\tau$ and Spearman's $\rho$ correlations for model pairs on the word similarity task. Section~\ref{sample_tab} gives examples of generated sentences for anchor words in contextualized evaluation. Section~\ref{prompting} describes the prompting design for generating samples, and Section~\ref{cos_figs} presents the cosine similarity distribution across all evaluation metrics.

\newpage

\input{Tables/Bats_performances}

\onecolumn
\subsubsection{Word Analogy Task Ranking}\label{analogy_ranking_apx}

\begin{figure*}[!h]\large
    \centering
    \includegraphics[width=\textwidth, height=\textheight,keepaspectratio]{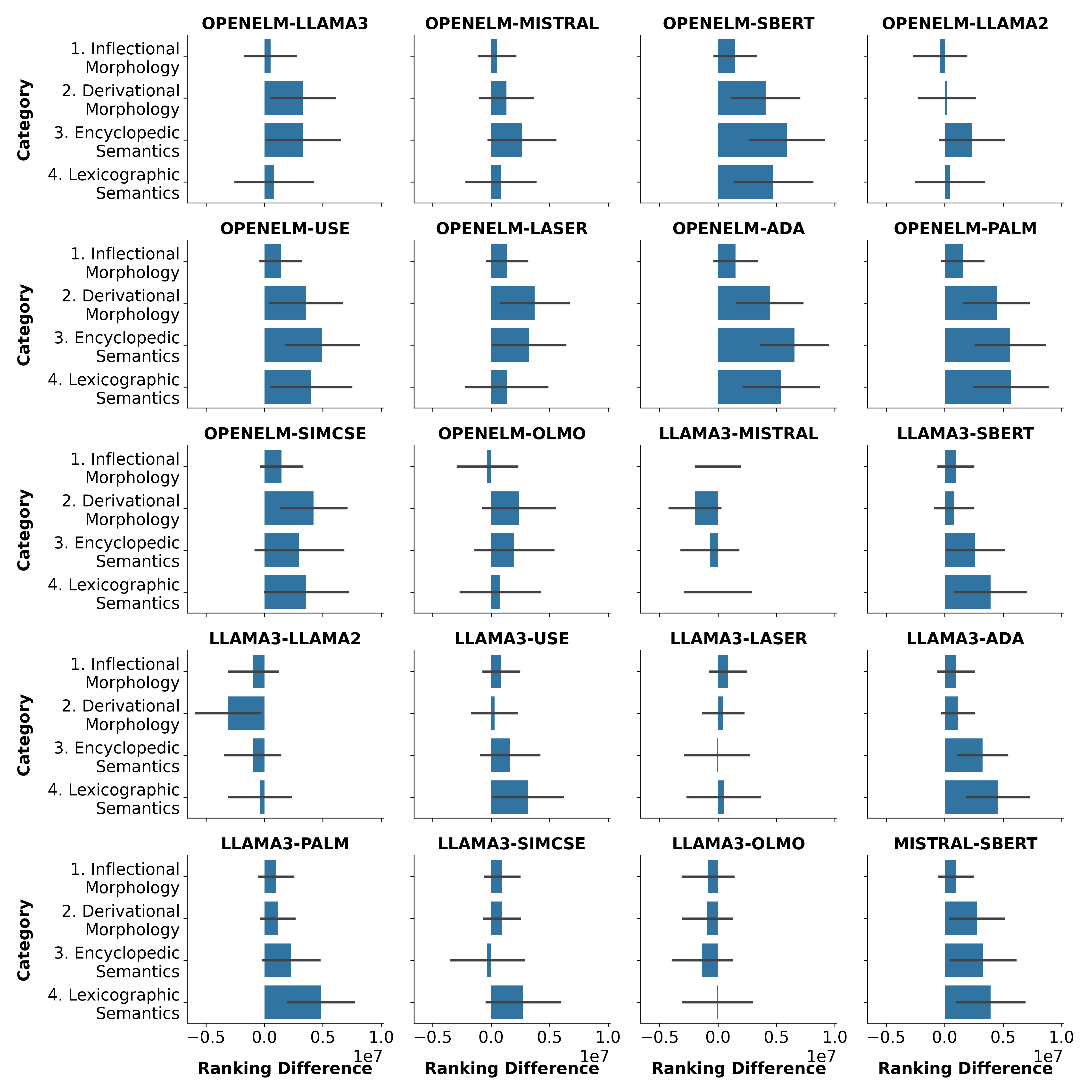}
    \caption{For each model, the cosine similarity of related words was found and ranked according to all pairs of words. Here, the difference in ranking between model pairs for certain BATS categories is shown.(\textit{Continued})}
    \label{fig:model-agreement-apx}
\end{figure*}

\begin{figure*}\ContinuedFloat
    \centering
    \includegraphics[width=\textwidth, height=\textheight,keepaspectratio]{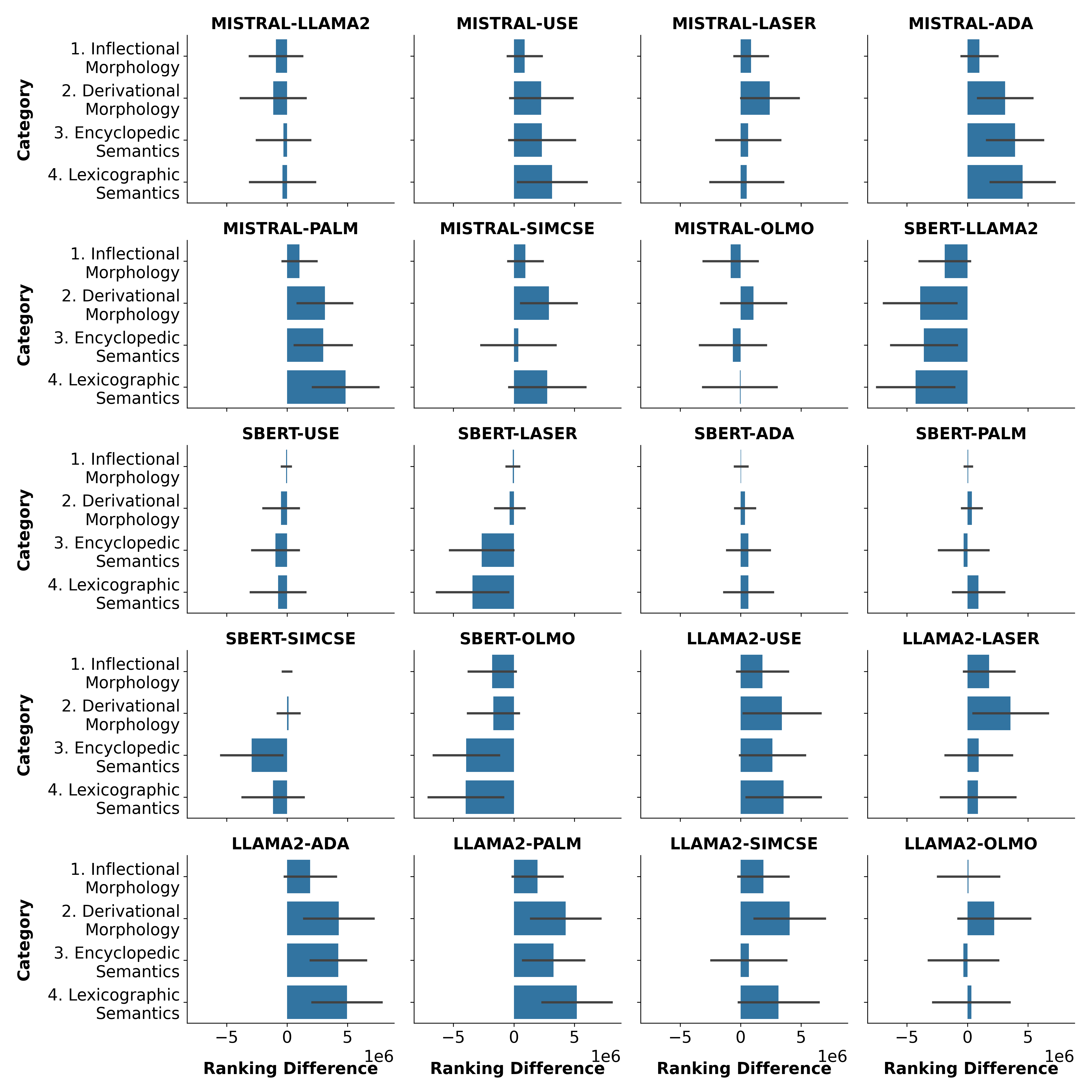}
    \caption{For each model, the cosine similarity of related words was found and ranked according to all pairs of words. Here, the difference in ranking between model pairs for certain BATS categories is shown. (\textit{Continued})}
    % \label{fig:model-agreement-b}
\end{figure*}

\begin{figure*}\ContinuedFloat
    \centering
    \includegraphics[width=\textwidth, height=\textheight,keepaspectratio]{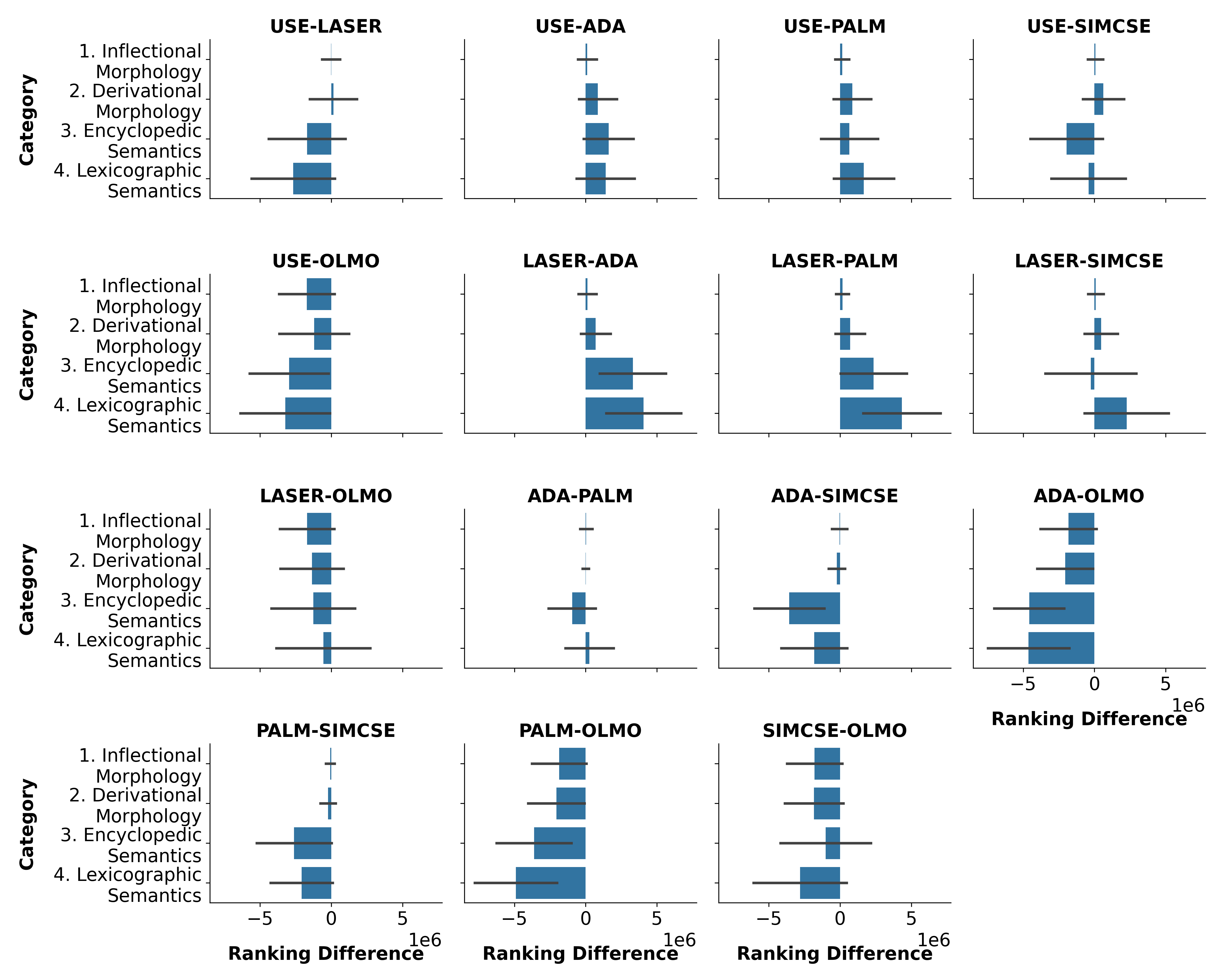}
    \caption{For each model, the cosine similarity of related words was found and ranked according to all pairs of words. Here, the difference in ranking between model pairs for certain BATS categories is shown.}
    % \label{fig:model-agreement-c}
\end{figure*}

% \begin{figure*}[!htb]
%     \centering
%     \includegraphics[width=\linewidth]{images/rankings1.pdf}
%     \vspace{-2mm}
%     \caption{For each model, the cosine similarity of related words was found and ranked according to all pairs of words. Here, the difference in ranking between model pairs for certain BATS categories is shown. }
%     \label{fig:model-agreement}
%     \vspace{-5mm}
% \end{figure*}

% \begin{figure*}[!htb]\ContinuedFloat
%     \centering
%     \includegraphics[width=\linewidth]{images/rankings2.pdf}
%     \vspace{-2mm}
%     \caption*{Figure~\ref{fig:model-agreement}: For each model, the cosine similarity of related words was found and ranked according to all pairs of words. Here, the difference in ranking between model pairs for certain BATS categories is shown.}
%     % \label{fig:model-agreement}
%     \vspace{-5mm}
% \end{figure*}

\newpage
\subsection{Similarity Correlation Analysis}\label{sim_corr}

\begin{figure*}[!htb]
     \centering
     \begin{subfigure}[b]{0.6\linewidth}
         \centering
         \includegraphics[width=\linewidth]{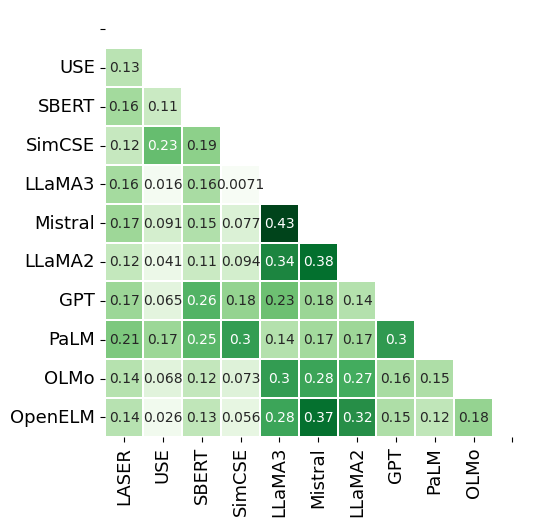}
         \caption{Kendal $\tau$}
         
     \end{subfigure}
     ~
     \begin{subfigure}[b]{0.6\linewidth}
         \centering
         \includegraphics[width=\linewidth]{new_plots/Spearman.png}
         \caption{Spearman $\rho$}
         
     \end{subfigure}
    
    \caption{Correlation coefficients for each pair of models, found using a large dataset of pairs of words.}
    \label{fig:model-corr-coef_apx}
    
\end{figure*}

\section{Contextualized Evaluation Setting}\label{sample_tab}

\input{Tables/samples}

\twocolumn
\subsection{Synthetic Data Generation}\label{prompting}

\subsubsection{Questionnaire}
\input{prompting_box/que_prompt}

\subsubsection{Active-Passive}
\input{prompting_box/act_prompt}

\subsection{Polysemy}
\input{prompting_box/poly_prompt}

\subsubsection{Paraphrase}
\input{prompting_box/para_prompt}

\subsubsection{Jumbling}
\input{prompting_box/jumbling_prompt}

\subsubsection{Synonym}
\input{prompting_box/syn_prompt}

\subsubsection{Negation}
\input{prompting_box/neg_prompt}

\subsubsection{Antonym}
\input{prompting_box/anto_prompt}

\section{Comparison of Models in Contextualized Settings}\label{cos_figs}
\begin{figure*}
    \centering
    \begin{subfigure}[b]{\textwidth}
        \includegraphics[width=\columnwidth]{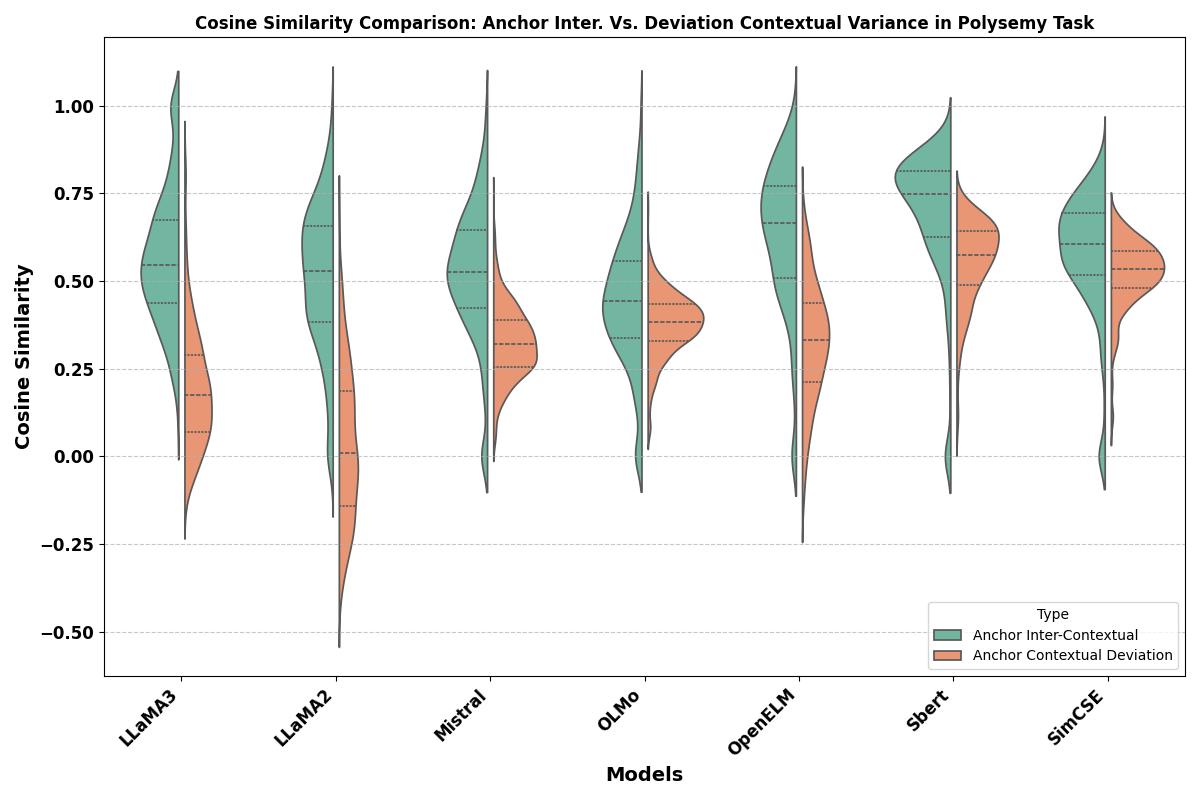}
        \caption{The distribution of cosine similarities between Anchor Inter-Contextual Variance and Anchor Contextual Deviation words.}
    \end{subfigure}
    \begin{subfigure}[b]{\textwidth}
        \includegraphics[width=\columnwidth]{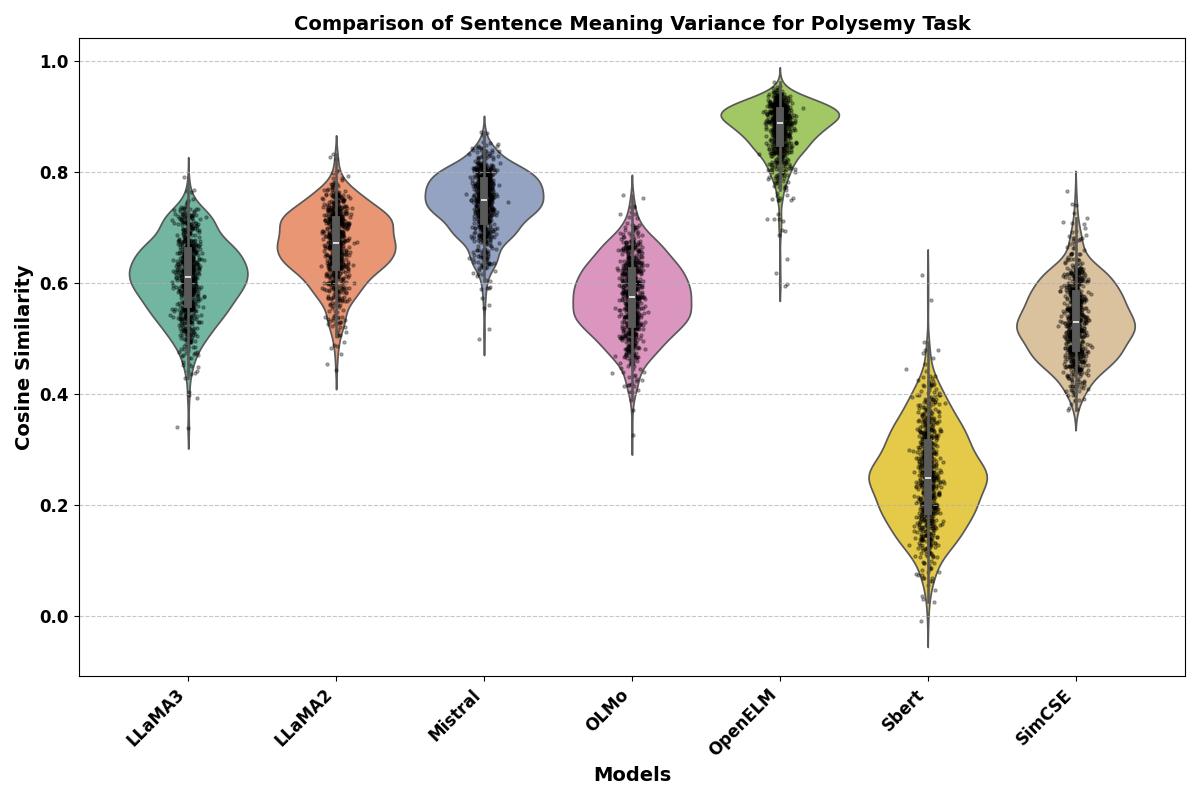}
        \caption{The distribution of cosine similarities between sentences in Sentence Meaning Variance.}
    
    \end{subfigure}
    \caption{Polysemy Task comparison}
    \label{fig:poly_plot}
\end{figure*}

\begin{figure*}[!ht]\small
\centering
\begin{subfigure}[b]{\textwidth}
    \includegraphics[width=\columnwidth]{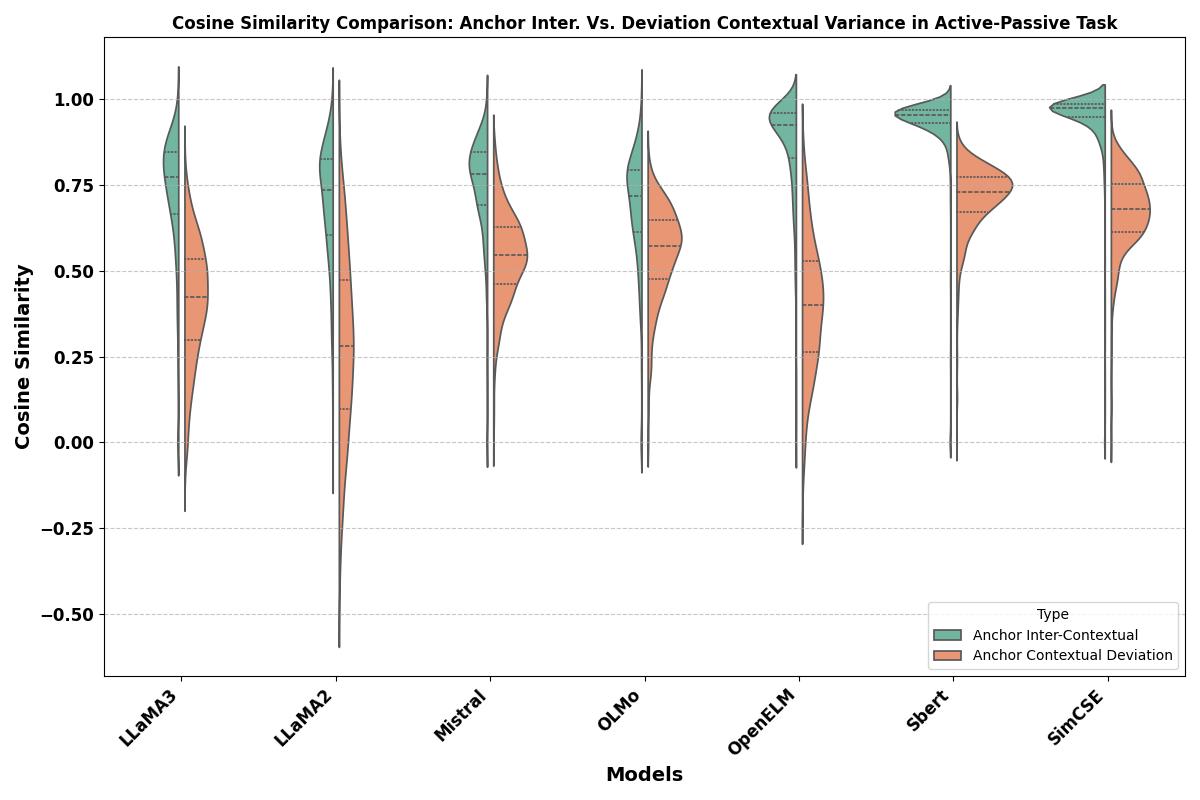}
    \caption{The distribution of cosine similarities between Anchor Inter-Contextual Variance and Anchor Contextual Deviation words. }
    
    \end{subfigure}
\begin{subfigure}[b]{\textwidth}
    \includegraphics[width=\columnwidth]{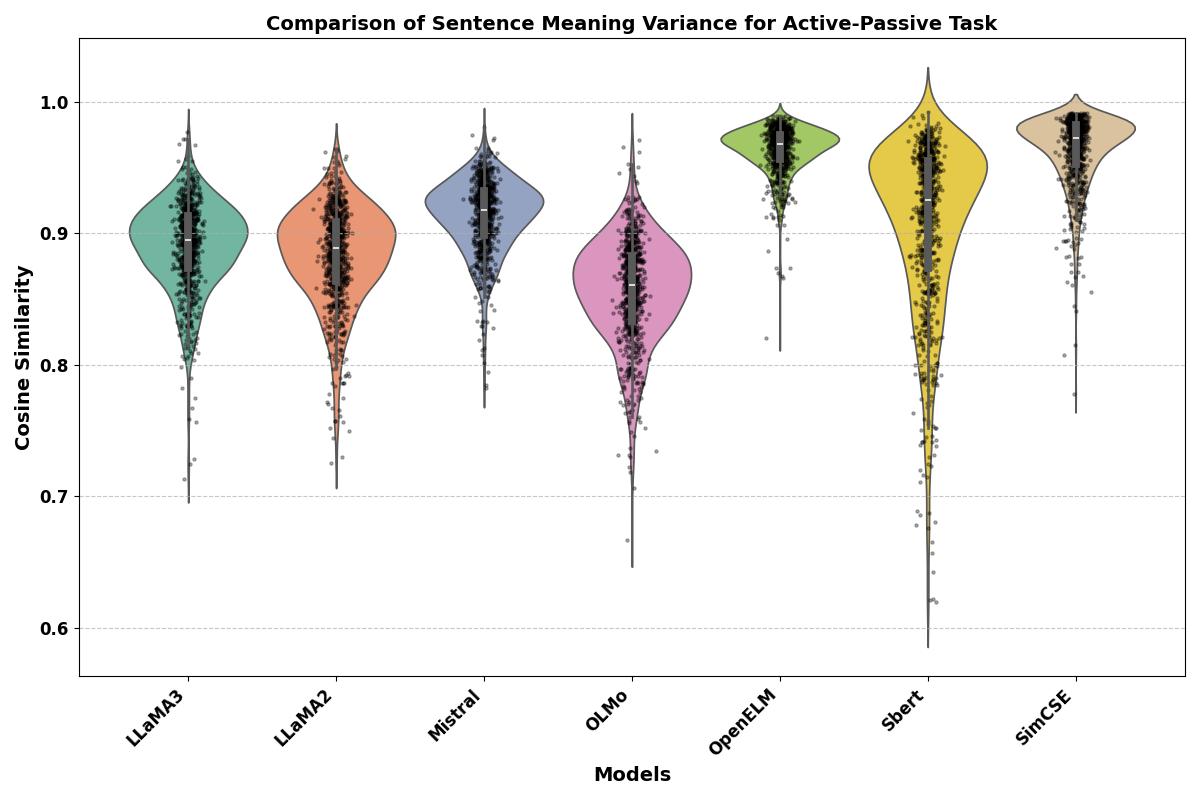}
    \caption{The distribution of cosine similarities between sentences in Sentence Meaning Variance.}
    
    \end{subfigure}
    \caption{Active-Passive Task comparison}
    \label{active_passive_plot}
\end{figure*}

\begin{figure*}[!ht]\small
\centering
\begin{subfigure}[b]{\textwidth}
    \includegraphics[width=\columnwidth]{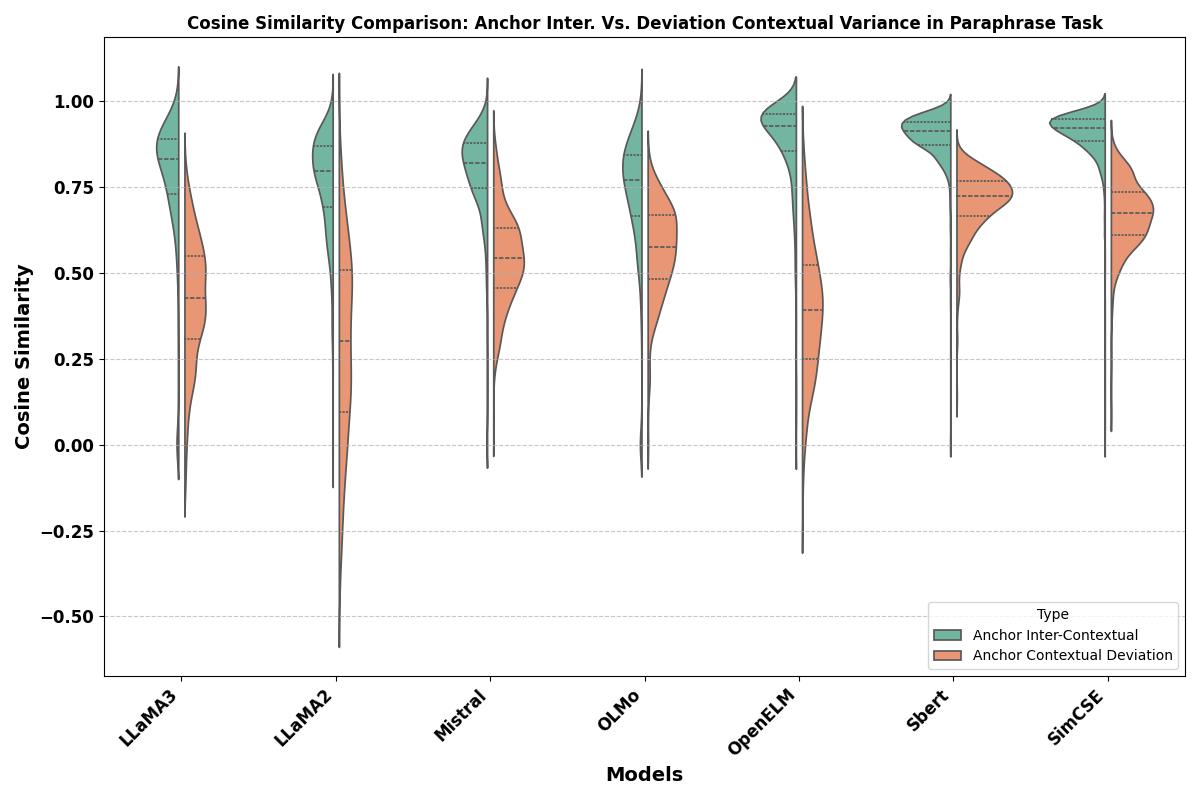}
    \caption{The distribution of cosine similarities between Anchor Inter-Contextual Variance and Anchor Contextual Deviation words.}
    
    \end{subfigure}
\begin{subfigure}[b]{\textwidth}
    \includegraphics[width=\columnwidth]{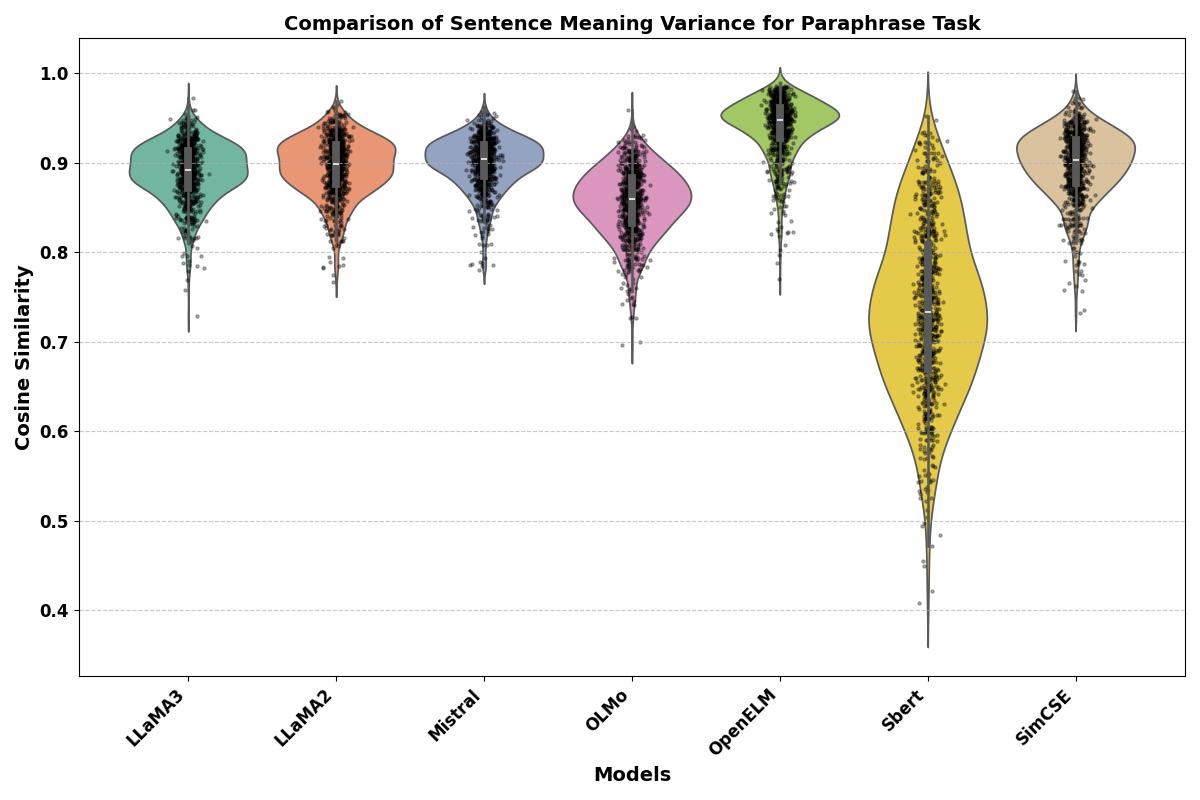}
    \caption{The distribution of cosine similarities between sentences in Sentence Meaning Variance.}
    
    \end{subfigure}
    \caption{Paraphrase Task comparison}
    \label{para_plots}
\end{figure*}

\begin{figure*}[!ht]\small
\centering
\begin{subfigure}[b]{\textwidth}
    \includegraphics[width=\columnwidth]{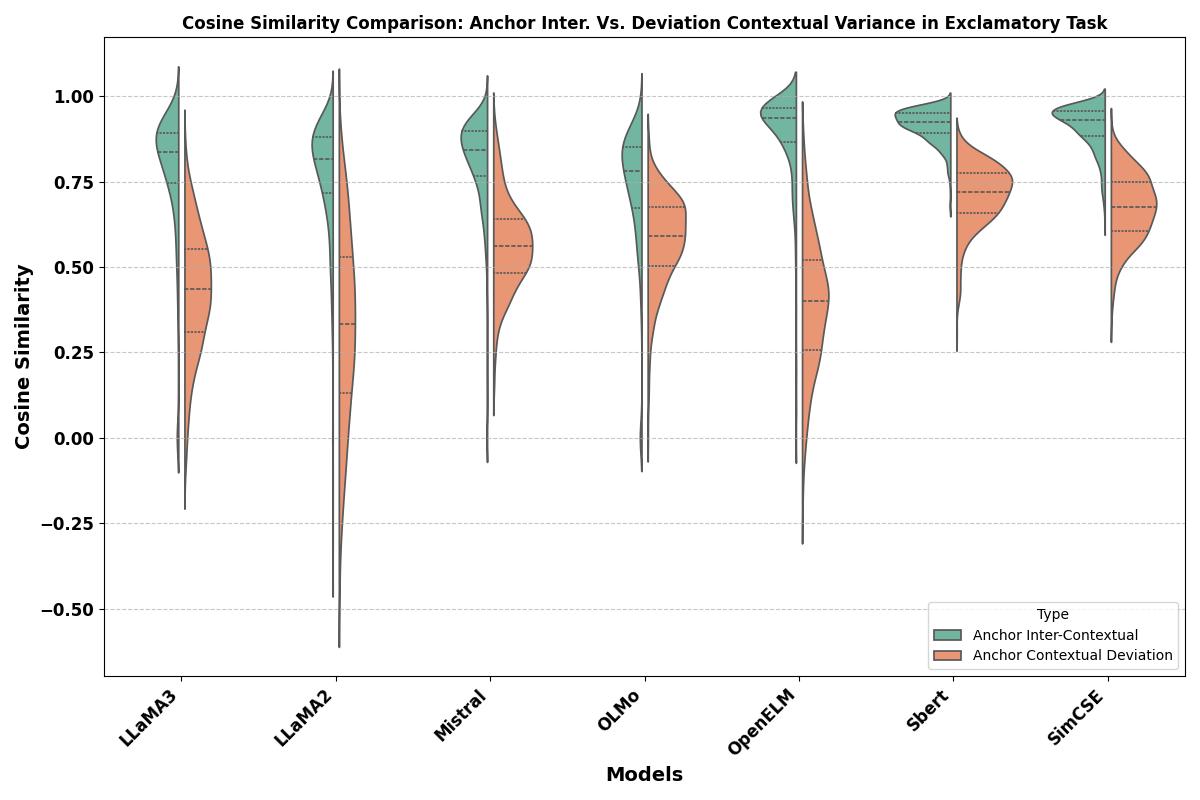}
    \caption{The distribution of cosine similarities between Anchor Inter-Contextual Variance and Anchor Contextual Deviation words.}
    
    \end{subfigure}
\begin{subfigure}[b]{\textwidth}
    \includegraphics[width=\columnwidth]{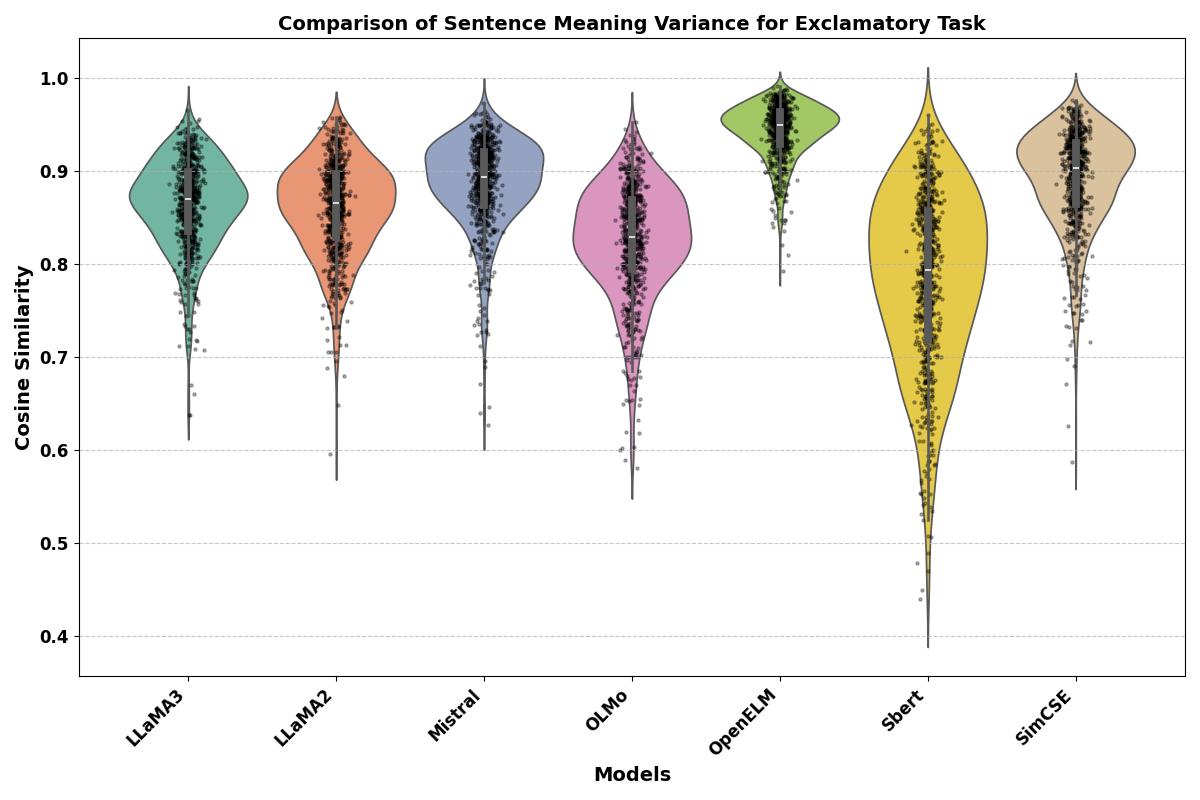}
    \caption{The distribution of cosine similarities between sentences in Sentence Meaning Variance.}
    
    \end{subfigure}
    \caption{Exclamatory Task comparison}
    \label{exclamatory_plots}
\end{figure*}

\begin{figure*}[!ht]\small
\centering
\begin{subfigure}[b]{\textwidth}
    \includegraphics[width=\columnwidth]{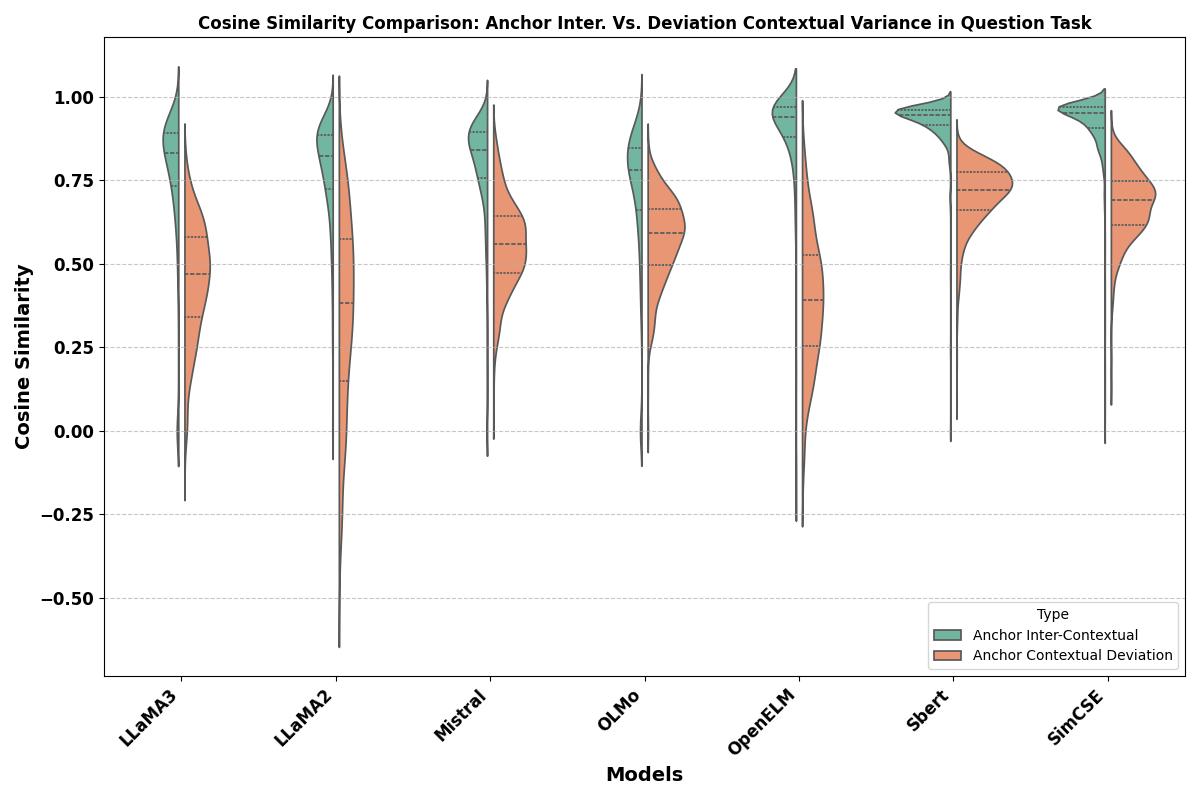}
    \caption{The distribution of cosine similarities between Anchor Inter-Contextual Variance and Anchor Contextual Deviation words.}
    
    \end{subfigure}
\begin{subfigure}[b]{\textwidth}
    \includegraphics[width=\columnwidth]{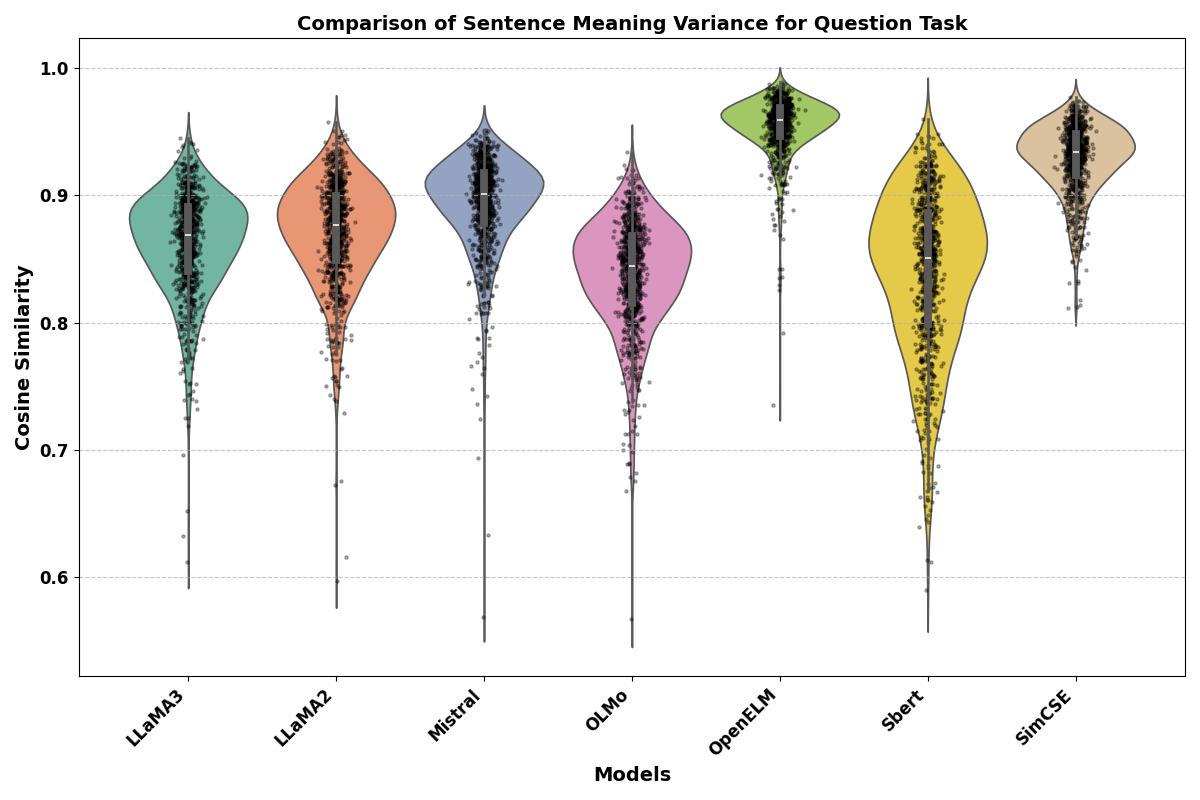}
    \caption{The distribution of cosine similarities between sentences in Sentence Meaning Variance.}
    
    \end{subfigure}
    \caption{Questionnaire Task comparison}
    \label{question_plots}
\end{figure*}

\begin{figure*}[!ht]\small
\centering
\begin{subfigure}[b]{\textwidth}
    \includegraphics[width=\columnwidth]{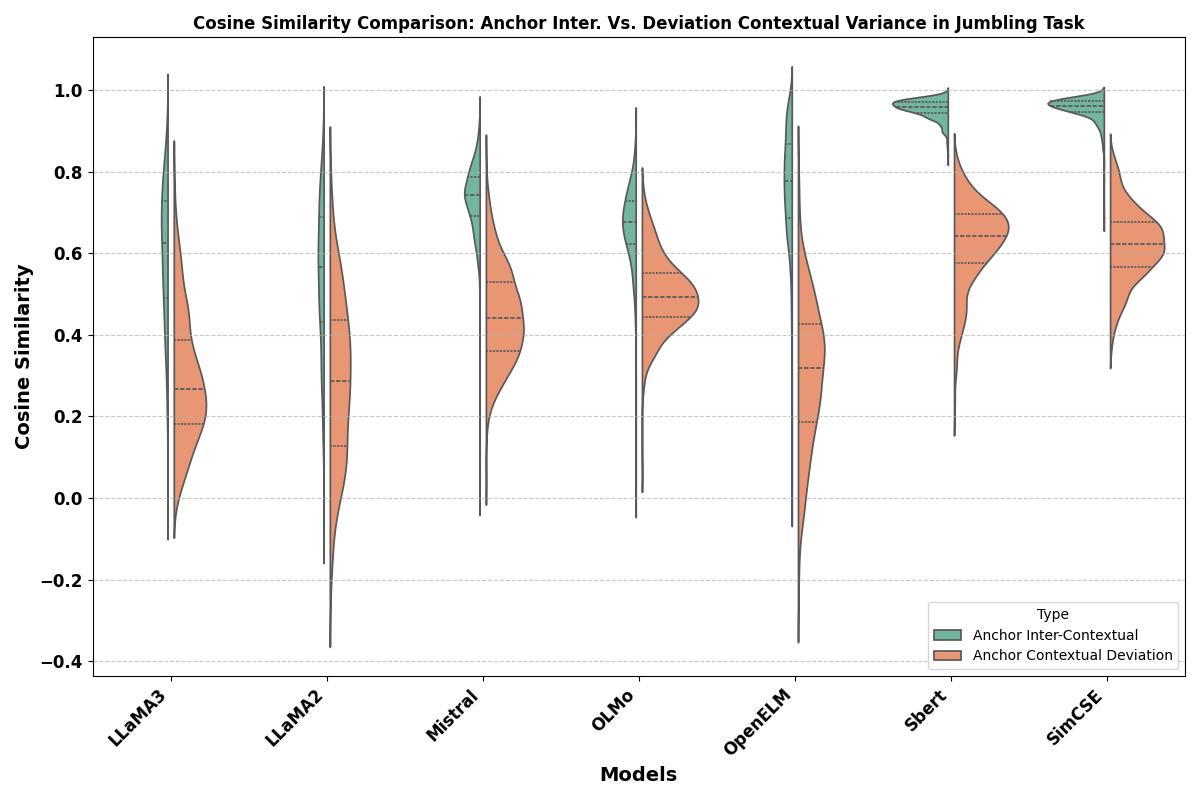}
    \caption{The distribution of cosine similarities between Anchor Inter-Contextual Variance and Anchor Contextual Deviation words.}
    
    \end{subfigure}
\begin{subfigure}[b]{\textwidth}
    \includegraphics[width=\columnwidth]{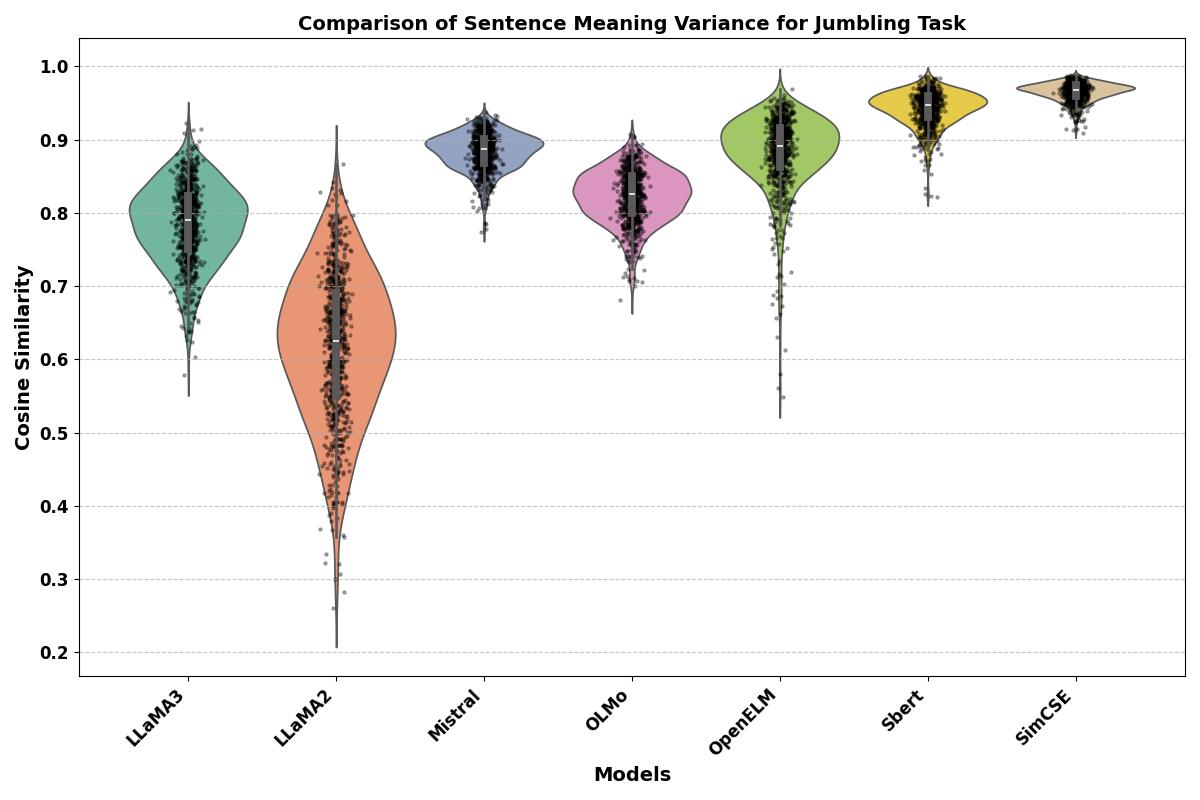}
    \caption{The distribution of cosine similarities between sentences in Sentence Meaning Variance.}
    
    \end{subfigure}
    \caption{Jumbling Task comparison}
    \label{jumbling_plots}
\end{figure*}

\begin{figure*}[!ht]\small
\centering
\begin{subfigure}[b]{\textwidth}
    \includegraphics[width=\columnwidth]{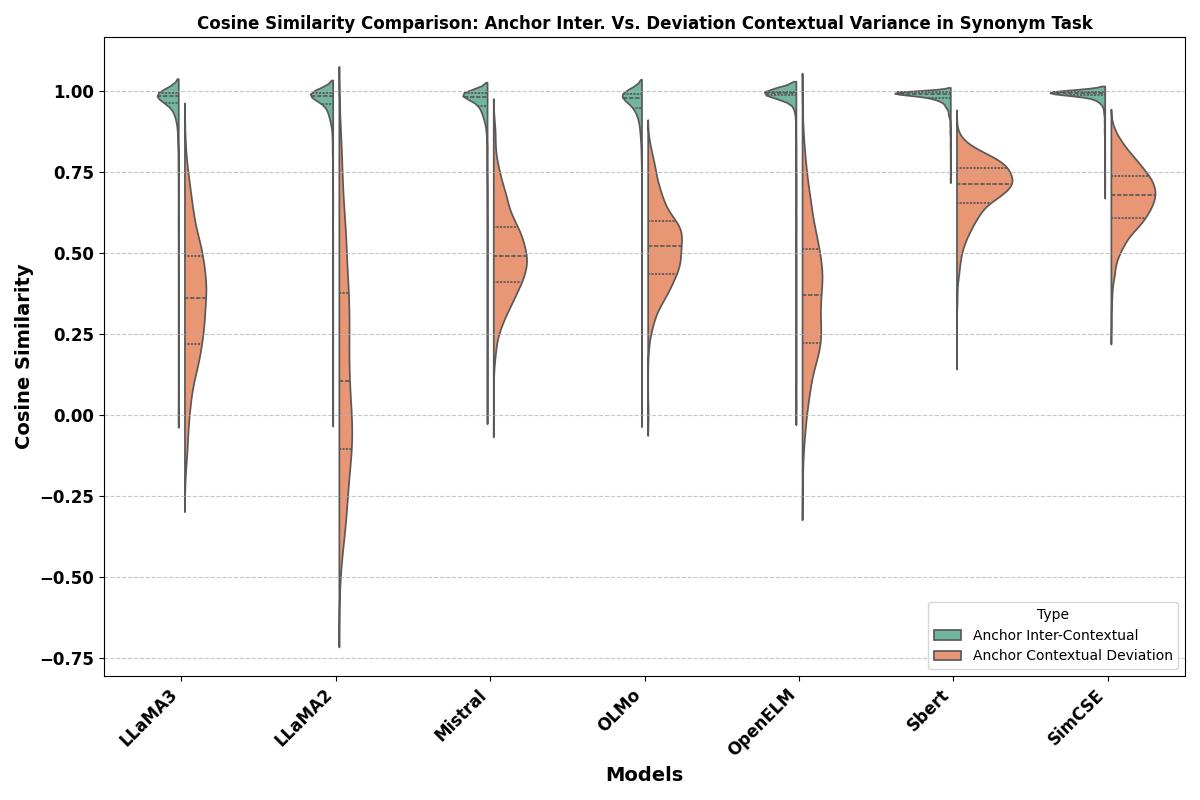}
    \caption{The distribution of cosine similarities between Anchor Inter-Contextual Variance and Anchor Contextual Deviation words.}
    \end{subfigure}
\begin{subfigure}[b]{\textwidth}
    \includegraphics[width=\columnwidth]{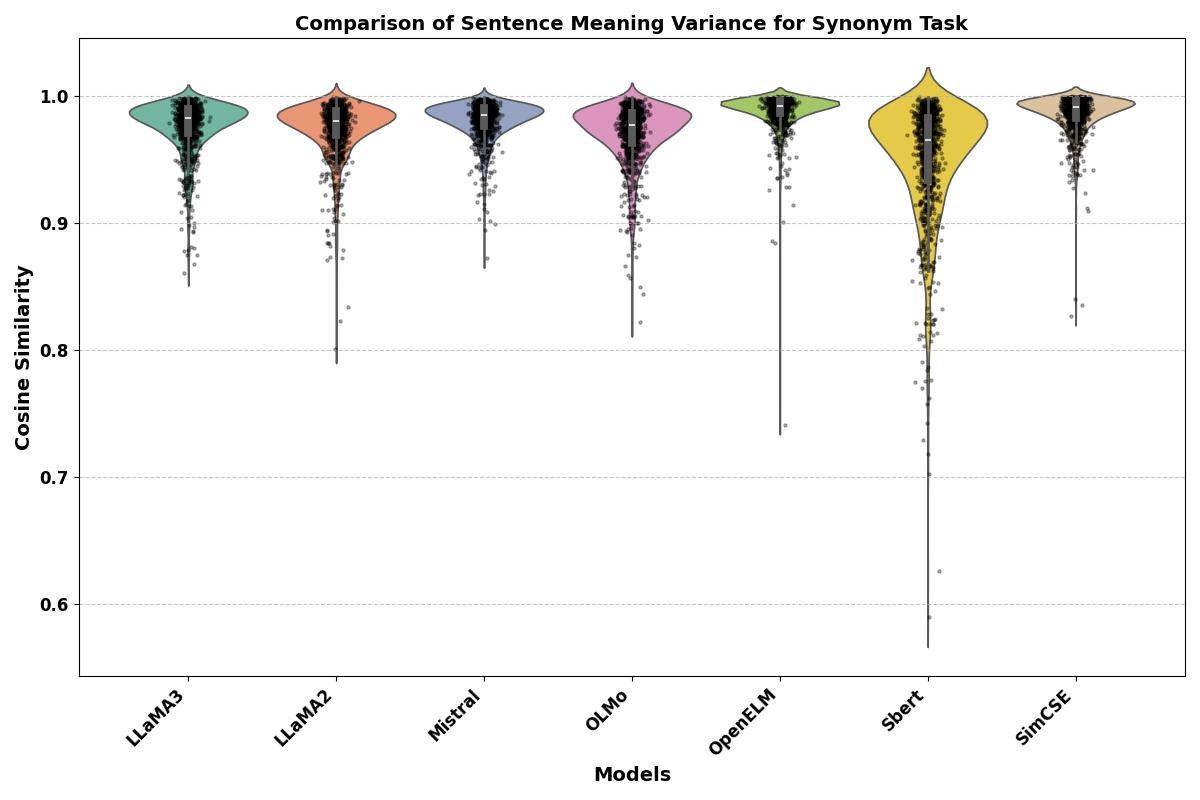}
    \caption{The distribution of cosine similarities between sentences in Sentence Meaning Variance.}
    \end{subfigure}
    \caption{Synonym Task comparison}
    \label{syn_plots}
\end{figure*}

\begin{figure*}[!ht]\small
\centering
\begin{subfigure}[b]{\textwidth}
    \includegraphics[width=\columnwidth]{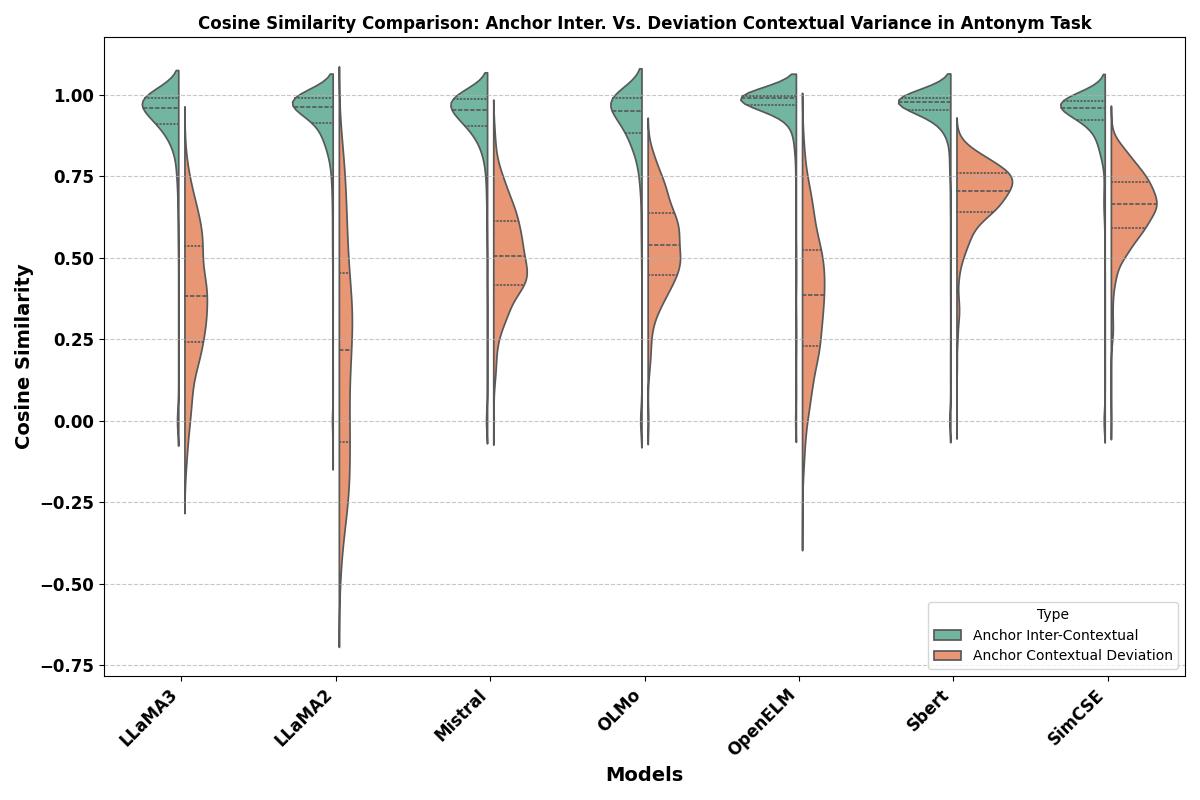}
    \caption{The distribution of cosine similarities between Anchor Inter-Contextual Variance and Anchor Contextual Deviation words.}
    \end{subfigure}
\begin{subfigure}[b]{\textwidth}
    \includegraphics[width=\columnwidth]{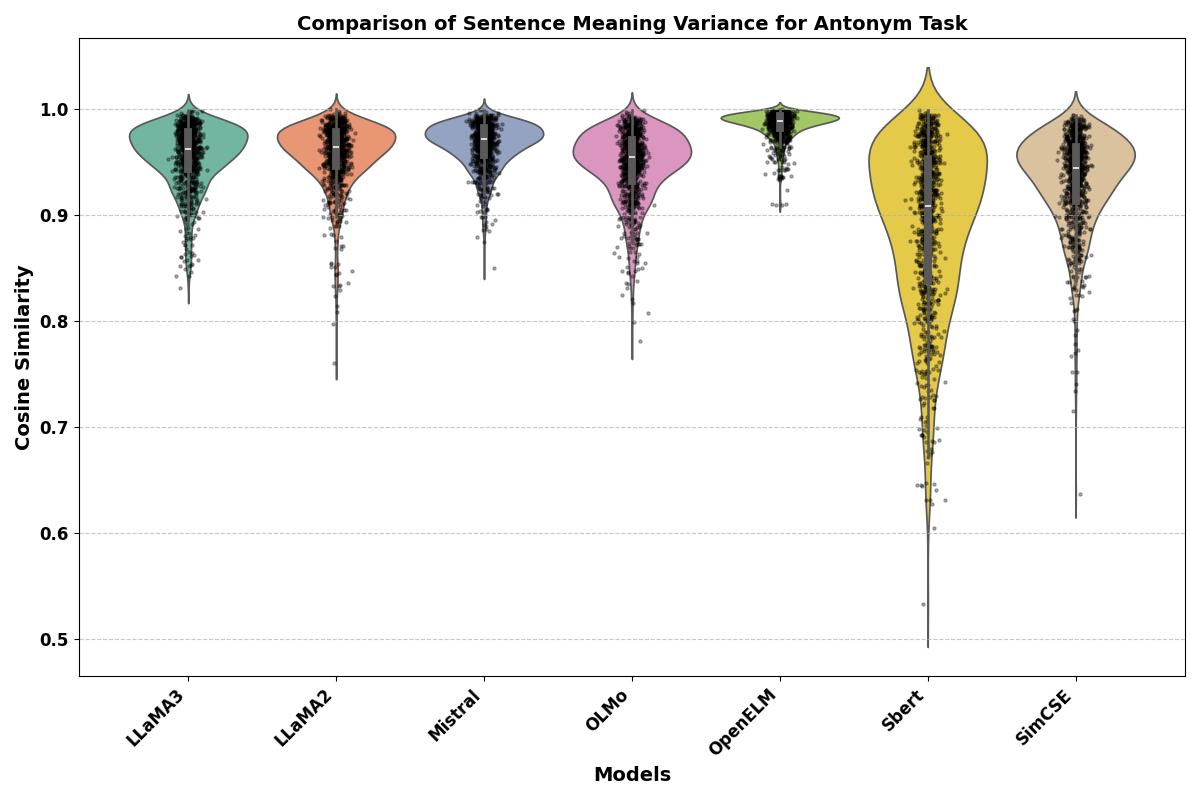}
    \caption{The distribution of cosine similarities between sentences in Sentence Meaning Variance.}
    \end{subfigure}
    \caption{Antonym Task comparison}
    \label{anto_plots}
\end{figure*}

\begin{figure*}[!ht]\small
\centering
\begin{subfigure}[b]{\textwidth}
    \includegraphics[width=\columnwidth]{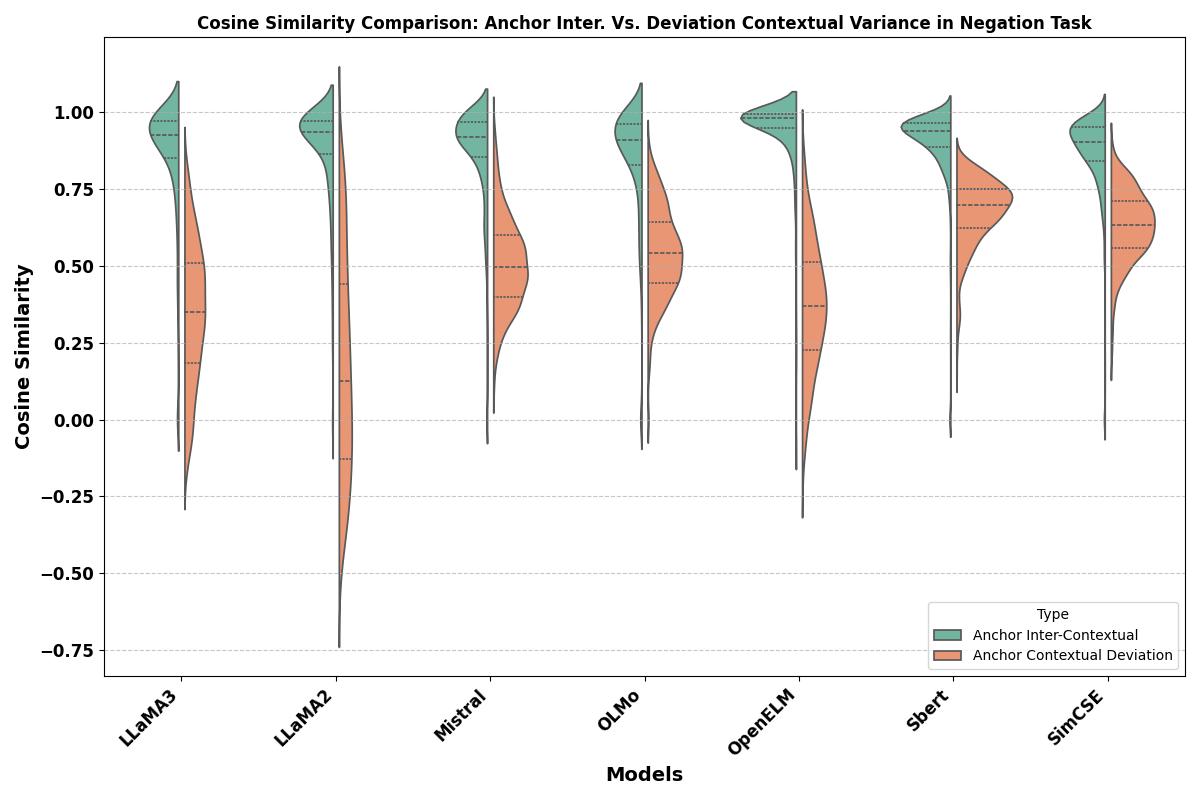}
    \caption{The distribution of cosine similarities between Anchor Inter-Contextual Variance and Anchor Contextual Deviation words.}
    
    \end{subfigure}
\begin{subfigure}[b]{\textwidth}
    \includegraphics[width=\columnwidth]{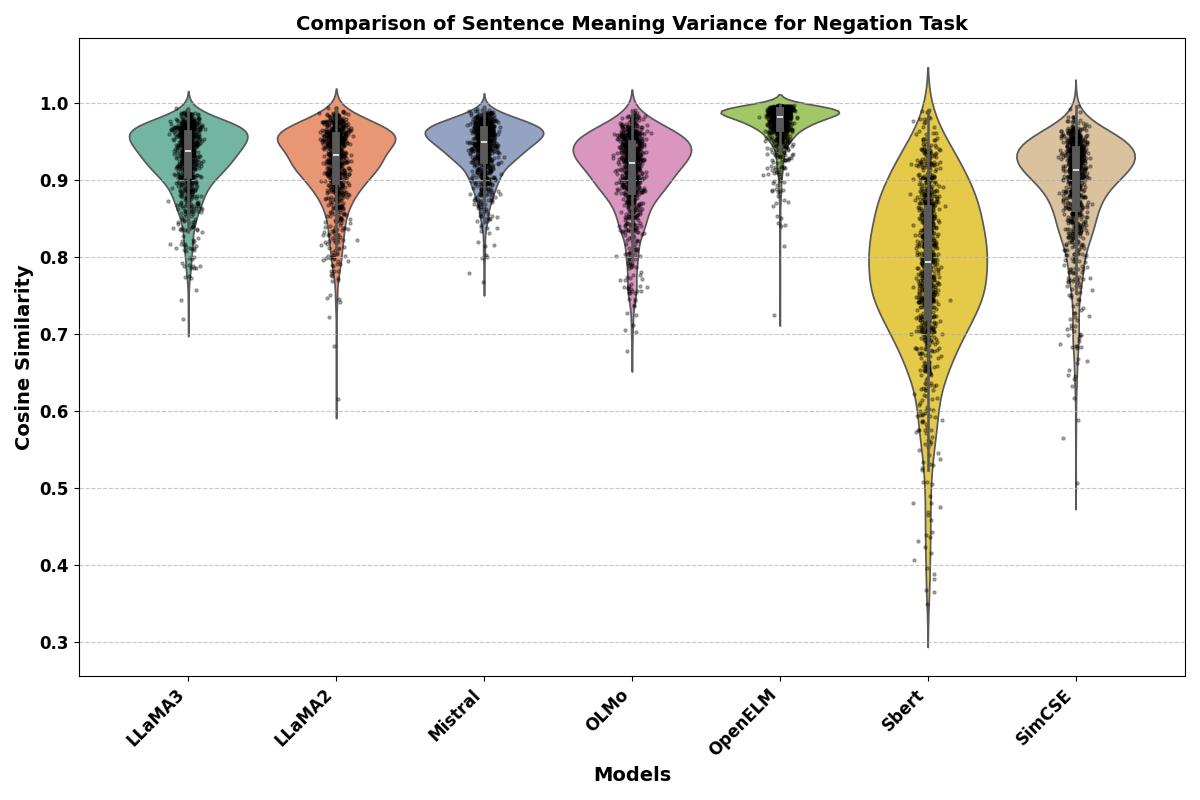}
    \caption{The distribution of cosine similarities between sentences in Sentence Meaning Variance.}
    
    \end{subfigure}
    \caption{Negation Task comparison}
    \label{neg_plots}
\end{figure*}

%% file: Tables/Bats_performances.tex
\sisetup{
  round-mode = places,  % Rounds numbers to the specified number of places
  round-precision = 2   % Number of decimal places to round to
}

\onecolumn

\begin{longtable}{c|llcccc}

\toprule
 \textbf{Model} & \textbf{Analogy Method} & \thead[cc]{1. Inflectional\\Morphology} & \thead[cc]{2. Derivational\\Morphology} & \thead[cc]{3. Encyclopedic\\Semantics} & \thead[cc]{4. Lexicographic\\Semantics} \\
\midrule
\endhead

\multirow{7}{*}{\textbf{GPT3-Ada}} 
& 3CosAdd & 0.761 & 0.677 & 0.115 & 0.097 \\
 & 3CosAvg & 0.802 & 0.734 & 0.148 & 0.102 \\
 & 3CosMul & 0.776 & 0.697 & 0.122 & 0.100 \\
 & LRCos & 0.606 & 0.482 & 0.280 & 0.132 \\
 & PairDistance & 0.546 & 0.323 & 0.052 & 0.006 \\
 & SimilarToAny & 0.155 & 0.044 & 0.005 & 0.029 \\
 & SimilarToB & 0.276 & 0.134 & 0.038 & 0.090 \\
\cline{1-6}
\multirow{7}{*}{\textbf{LLaMA2}} 
& 3CosAdd & 0.230 & 0.271 & 0.055 & 0.023 \\
 & 3CosAvg & 0.326 & 0.362 & 0.086 & 0.026 \\
 & 3CosMul & 0.230 & 0.276 & 0.053 & 0.022 \\
 & LRCos & 0.150 & 0.148 & 0.176 & 0.050 \\
 & PairDistance & 0.066 & 0.130 & 0.013 & 0.001 \\
 & SimilarToAny & 0.065 & 0.043 & 0.037 & 0.011 \\
 & SimilarToB & 0.130 & 0.118 & 0.054 & 0.026 \\
\cline{1-6}
\multirow{7}{*}{\textbf{LLaMA3}} 
& 3CosAdd & 0.079 & 0.099 & 0.011 & 0.009 \\
 & 3CosAvg & 0.096 & 0.114 & 0.016 & 0.010 \\
 & 3CosMul & 0.076 & 0.097 & 0.010 & 0.009 \\
 & LRCos & 0.044 & 0.058 & 0.104 & 0.006 \\
 & PairDistance & 0.001 & 0.004 & 0.000 & 0.002 \\
 & SimilarToAny & 0.053 & 0.059 & 0.010 & 0.008 \\
 & SimilarToB & 0.100 & 0.112 & 0.018 & 0.016 \\
\cline{1-6}
\multirow{7}{*}{\textbf{Mistral}} 
& 3CosAdd & 0.084 & 0.093 & 0.010 & 0.010 \\
 & 3CosAvg & 0.102 & 0.116 & 0.018 & 0.012 \\
 & 3CosMul & 0.082 & 0.090 & 0.010 & 0.009 \\
 & LRCos & 0.066 & 0.068 & 0.110 & 0.010 \\
 & PairDistance & 0.001 & 0.006 & 0.000 & 0.003 \\
 & SimilarToAny & 0.062 & 0.063 & 0.008 & 0.009 \\
 & SimilarToB & 0.108 & 0.112 & 0.014 & 0.012 \\
\cline{1-6}
\multirow{7}{*}{\textbf{OLMo}} 
& 3CosAdd & 0.094 & 0.093 & 0.014 & 0.009 \\
 & 3CosAvg & 0.116 & 0.106 & 0.022 & 0.014 \\
 & 3CosMul & 0.090 & 0.089 & 0.012 & 0.009 \\
 & LRCos & 0.074 & 0.078 & 0.100 & 0.014 \\
 & PairDistance & 0.001 & 0.004 & 0.000 & 0.002 \\
 & SimilarToAny & 0.065 & 0.057 & 0.012 & 0.008 \\
 & SimilarToB & 0.116 & 0.108 & 0.022 & 0.016 \\
\cline{1-6}
\multirow{7}{*}{\textbf{OpenELM}} 
& 3CosAdd & 0.030 & 0.031 & 0.003 & 0.004 \\
 & 3CosAvg & 0.070 & 0.052 & 0.010 & 0.008 \\
 & 3CosMul & 0.025 & 0.027 & 0.002 & 0.003 \\
 & LRCos & 0.002 & 0.002 & 0.046 & 0.004 \\
 & PairDistance & 0.003 & 0.003 & 0.000 & 0.002 \\
 & SimilarToAny & 0.040 & 0.035 & 0.007 & 0.005 \\
 & SimilarToB & 0.066 & 0.054 & 0.012 & 0.008 \\
\cline{1-6}

\multirow{7}{*}{\textbf{PaLM}} & 3CosAdd & 0.743 & 0.609 & 0.118 & 0.122 \\
 & 3CosAvg & 0.794 & 0.668 & 0.232 & 0.136 \\
 & 3CosMul & 0.768 & 0.648 & 0.128 & 0.124 \\
 & LRCos & 0.780 & 0.714 & 0.404 & 0.238 \\
 & PairDistance & 0.466 & 0.249 & 0.048 & 0.008 \\
 & SimilarToAny & 0.165 & 0.027 & 0.011 & 0.035 \\
 & SimilarToB & 0.270 & 0.082 & 0.030 & 0.108 \\
\cline{1-6}
\multirow{7}{*}{\textbf{SBERT}} & 3CosAdd & 0.461 & 0.393 & 0.046 & 0.073 \\
 & 3CosAvg & 0.474 & 0.418 & 0.058 & 0.092 \\
 & 3CosMul & 0.506 & 0.424 & 0.062 & 0.074 \\
 & LRCos & 0.808 & 0.642 & 0.270 & 0.228 \\
 & PairDistance & 0.135 & 0.184 & 0.021 & 0.003 \\
 & SimilarToAny & 0.178 & 0.065 & 0.003 & 0.019 \\
 & SimilarToB & 0.302 & 0.154 & 0.020 & 0.088 \\
\cline{1-6}
\multirow{7}{*}{\textbf{SimCSE}} & 3CosAdd & 0.040 & 0.045 & 0.008 & 0.007 \\
 & 3CosAvg & 0.058 & 0.068 & 0.016 & 0.012 \\
 & 3CosMul & 0.035 & 0.039 & 0.007 & 0.006 \\
 & LRCos & 0.024 & 0.026 & 0.070 & 0.006 \\
 & PairDistance & 0.001 & 0.002 & 0.001 & 0.002 \\
 & SimilarToAny & 0.036 & 0.037 & 0.010 & 0.007 \\
 & SimilarToB & 0.056 & 0.068 & 0.014 & 0.012 \\
\cline{1-6}
\multirow{7}{*}{\textbf{USE}} & 3CosAdd & 0.397 & 0.156 & 0.039 & 0.103 \\
 & 3CosAvg & 0.442 & 0.190 & 0.084 & 0.132 \\
 & 3CosMul & 0.436 & 0.165 & 0.049 & 0.100 \\
 & LRCos & 0.722 & 0.412 & 0.396 & 0.270 \\
 & PairDistance & 0.076 & 0.012 & 0.008 & 0.005 \\
 & SimilarToAny & 0.101 & 0.032 & 0.006 & 0.035 \\
 & SimilarToB & 0.204 & 0.098 & 0.026 & 0.098 \\
\cline{1-6}
\multirow{7}{*}{\textbf{LASER}} 
& 3CosAdd & 0.431 & 0.434 & 0.022 & 0.022 \\
 & 3CosAvg & 0.484 & 0.506 & 0.030 & 0.020 \\
 & 3CosMul & 0.448 & 0.454 & 0.023 & 0.023 \\
 & LRCos & 0.510 & 0.482 & 0.116 & 0.028 \\
 & PairDistance & 0.230 & 0.245 & 0.009 & 0.003 \\
 & SimilarToAny & 0.087 & 0.027 & 0.004 & 0.007 \\
 & SimilarToB & 0.198 & 0.072 & 0.012 & 0.020 \\
\cline{1-6}\multirow{7}{*}{\textbf{GloVe}}
&3CosAdd & 0.720 & 0.351 & 0.262 & 0.060 \\
&3CosAvg & 0.764 & 0.446 &0.430 & 0.076 \\
&3CosMul & 0.770 & 0.366 &0.228 & 0.017 \\
&LRCos  & 0.880 & 0.544 &0.440 & 0.086 \\
&PairDistance & 0.395 & 0.089 &0.122 & 0.003 \\
&SimilarToAny & 0.233 & 0.059 &0.089 & 0.051 \\
&SimilarToB  & 0.324 & 0.124 &0.132 & 0.062 \\
\cline{1-6}
\multirow{7}{*}{\textbf{Word2Vec}}
& 3CosAdd        & 0.775  & 0.319  & 0.137  & 0.062  \\
& 3CosAvg        & 0.828  & 0.376  & 0.266  & 0.072  \\
& 3CosMul        & 0.804  & 0.329  & 0.092  & 0.014  \\
& LRCos          & 0.932  & 0.600  & 0.224  & 0.086  \\
& PairDistance   & 0.355  & 0.054  & 0.070  & 0.003  \\
& SimilarToAny   & 0.254  & 0.094  & 0.074  & 0.052  \\
& SimilarToB     & 0.394  & 0.196  & 0.068  & 0.066  \\ 
\bottomrule
    \caption{BATS performance across categories with methods.}
    \label{tab:bats_analogy_performance}
\end{longtable}

% Switch back to two-column mode
\twocolumn

%% file: Tables/samples.tex
\begin{table*}[h]
\centering
\begin{tabular}{c|lp{10cm}}
\toprule
\textbf{Task} & \textbf{} & \textbf{Examples} \\
\midrule
\multirow{6}{*}{\textbf{Synonym}} & 
\vspace{1mm}
\textbf{Sentence-1}: &The actress was \textit{deeply} adored by her fans for her talent and humility. \\ 
% \cline{2-3}
\vspace{1mm}
& \textbf{Sentence-2}: &The actress was \textit{profoundly} adored by her fans for her talent and humility. \\ 
& \textbf{Word Replaced}: &\textit{deeply} \\ 
& \textbf{Word Replaced with}: &\textit{profoundly} \\ 
& \textbf{Anchor Word}: & \textbf{adored} \\
\midrule

% highlight the words with color for better readability.
\multirow{6}{*}{\textbf{Antonym}} & \textbf{Sentence-1}: 
\vspace{1mm}
&The brilliant sunset over the ocean was a sight everyone on the beach deeply \textit{cherished} and \textbf{adored}. \\
\vspace{1mm}
& \textbf{Sentence-2}: &The brilliant sunset over the ocean was a sight everyone on the beach deeply \textit{despised} and \textbf{adored}. \\
& \textbf{Word Replaced}: &\textit{cherished} \\
& \textbf{Word Replaced with}: &\textit{despised} \\
& \textbf{Anchor Word}: &\textbf{adored} \\
\midrule

\multirow{5}{*}{\textbf{Negation}} 
\vspace{1mm}
& \textbf{Sentence-1}: &The famous musician was \textbf{adored} by millions of fans worldwide. \\
\vspace{1mm}
& \textbf{Sentence-2}: &The famous musician was \textbf{\textit{not} adored} by millions of fans worldwide.\\
& \textbf{Anchor Word}: & \textbf{adored} \\
& \textbf{Negation Added}: & \textbf{\textit{not} adored} \\
\midrule

\multirow{12}{*}{\textbf{Jumbling}} 
\vspace{1mm}
& \textbf{Sentence-1}: &The famous actor was \textbf{adored} by millions of fans worldwide for his charismatic performances on the silver screen. \\
\vspace{1mm}
& \textbf{Sentence-2}: &\textit{was the famous actor} \textbf{adored} by millions of fans worldwide for his charismatic performances on the silver screen. \\ 
\vspace{1mm}
& \textbf{Sentence-3}: & \textit{on millions performances for the was silver screen. his \textbf{adored} charismatic actor of the by fans famous worldwide}\\
\vspace{1mm}
& \textbf{Sentence-4}: & the famous worldwide was \textbf{adored} by millions of fans actor for his charismatic performances on the silver screen.\\
\vspace{1mm}
& \textbf{Sentence-5}: & the the charismatic was \textbf{adored} by millions of fans worldwide for his actor performances on famous silver screen.\\
& \textbf{Anchor Word}: & \textbf{adored} \\
\midrule

\multirow{10}{*}{\textbf{Active-Passive}} 
\vspace{1mm}
& \textbf{Sentence-1}: &The talented musician was \textbf{adored} by fans for her soulful performances. \\
\vspace{1mm}
& \textbf{Sentence-2}: &Fans \textbf{adored} the talented musician for her soulful performances. \\
\vspace{1mm}
&\textbf{Sentence-3}: &Soulful performances were what fans \textbf{adored} about the talented musician.\\
\vspace{1mm}
&\textbf{Sentence-4}:& The musician's soulful performances made her \textbf{adored} by countless fans.\\
\vspace{1mm}
&\textbf{Sentence-5}: &The talented musician was enthusiastically \textbf{adored} by fans for delivering soulful performances.\\
\vspace{1mm}
& \textbf{Anchor Word}: &\textbf{adored} \\
\midrule
\end{tabular}
\caption{Task Examples (Part 1) (\textit{Continued})}
% \label{tab:samples}
\end{table*}

% #####################

\begin{table*}[h]\ContinuedFloat
\centering
\begin{tabular}{c|lp{10cm}}
\toprule
\textbf{Task} & \textbf{} & \textbf{Examples} \\
\midrule
\multirow{13}{*}{\textbf{Paraphrase}} 

& \textbf{Sentence-1}: &The famous actor was \textbf{adored} by millions of fans worldwide for his charismatic performances on the silver screen. \\
% \vspace{1mm}
& \textbf{Sentence-2}: &Legions of admirers cherished the renowned celebrity, who was \textbf{adored} for his magnetic screen presence and captivating portrayals. \\
&\textbf{Sentence-3}: & The iconic star was \textbf{adored} by countless devotees for his spellbinding acting prowess and mesmerizing big screen appearances.\\
&\textbf{Sentence-4}: & Multitudes of enthusiasts revered the legendary performer, whose \textbf{adored} on-screen personas and enthralling acting talents left an indelible mark. \\
&\textbf{Sentence-5}: &The revered thespian was \textbf{adored} by a global fanbase for his captivating performances and charismatic screen presence that enthralled audiences worldwide.\\
& \textbf{Anchor Word}: & \textbf{adored} \\
\midrule

\multirow{9}{*}{\textbf{Questionnaire}} 
\vspace{1mm}
& \textbf{Sentence-1}: &The famous celebrity was \textbf{adored} by millions of fans worldwide. \\
\vspace{1mm}
& \textbf{Sentence-2}: &Was the famous celebrity \textbf{adored} by millions of fans worldwide? \\
\vspace{1mm}
&\textbf{Sentence-3}: &Did the famous celebrity was \textbf{adored} by millions of fans across the globe?\\
\vspace{1mm}
&\textbf{Sentence-4}: &Were there millions of fans worldwide who \textbf{adored} the famous celebrity?\\
\vspace{1mm}
&\textbf{Sentence-5}: &Has the famous celebrity been \textbf{adored} by a vast number of fans globally?\\
\vspace{1mm}
& \textbf{Anchor Word}: &\textbf{adored} \\
\midrule

\multirow{9}{*}{\textbf{Exclamation}} 
\vspace{1mm}
& \textbf{Sentence-1}: &The \textbf{adored} celebrity was swarmed by fans seeking autographs and selfies. \\
\vspace{1mm}
& \textbf{Sentence-2}: &How \textbf{adored} the celebrity was by the fans who swarmed them for autographs and selfies!\\
\vspace{1mm}
&\textbf{Sentence-3}: &What an \textbf{adored} celebrity, to be swarmed by so many fans seeking autographs and selfies!\\
\vspace{1mm}
&\textbf{Sentence-4}:& How the fans \textbf{adored} the celebrity, swarming them for autographs and selfies!\\
\vspace{1mm}
&\textbf{Sentence-5}:& \textbf{adored} beyond measure, the celebrity found themselves swarmed by fans - what a scene of autographs and selfies!\\
\vspace{1mm}
& \textbf{Anchor Word}: &\textbf{adored} \\
\midrule

\multirow{10}{*}{\textbf{Polysemic}} 
& \textbf{Sentence-1}: &The CEO delivered an inspiring \textbf{address} to the company employees during the annual meeting. \\
& \textbf{Sentence-2}: &Could you please provide me with your current residential \textbf{address} for our records?\\
&\textbf{Sentence-3}: &The computer program accessed the memory \textbf{address} to retrieve the data.\\
&\textbf{Sentence-4}: & The speaker began her \textbf{address} by thanking the audience for attending.\\
&\textbf{Sentence-5}: & Please \textbf{address} the envelope carefully to ensure it reaches the correct destination.\\
& \textbf{Anchor Word}: &\textbf{address} \\
\midrule
\end{tabular}
\caption{Task Examples (Part 2)}
\label{tab:samples}
\end{table*}

%% file: prompting_box/que_prompt.tex
\begin{boxI}\small
\label{que_prompt}
\textit{\textbf{Questionnaire Task Generation Prompt}}: \\\
\textbf{`System Prompt'}: \\
        Using the anchor word, create a sentence S1 that includes the anchor word. After generating S1, generate four more questionnaire sentences of S1. It's crucial that all sentences retain the anchor word in its original form in all sentences. \\
        
         Here is an example. For a given anchor word 'forum', the generated S1 and S2 sentences are:\\
        \{\\
        'sentence1': "The online forum provides a platform for experts to discuss emerging technologies.",\\
        'sentence2': "Does the online forum provide a platform for experts to discuss emerging technologies?",\\
        'anchor\_word': 'forum'\\
        \}\\
        
        The output should be in the following json format: \\
        \{'sentence1: S1,\\
        'sentence2': S2, \\
        'sentence3: S3,\\
        'sentence4': S4, \\
        'sentence5': S5, \\
        'anchor\_word': anchor\_word\\
        \}\\
        
        \textbf{User}: Here is the anchor word: {word}. Note that, The anchor word must appear unchanged in all sentences.
    
\end{boxI}

%% file: prompting_box/act_prompt.tex
\begin{boxI}\small
\label{act_prompt}
\textit{\textbf{Active-Passive Task Generation Prompt}}: \\\
\textbf{`System Prompt'}: \\
        Using the anchor word, create an active voice sentence S1 that includes the anchor word. After generating S1, generate four passive voice sentences of S1. It's crucial that all sentences retain the anchor word in its original form in all the sentences.\\
        
         Here is an example, for a given anchor word 'forum', the generated S1 and S2 sentences are:\\
        \{ 'sentence1': "Experts frequently share their knowledge in the online forum about emerging technologies.",\\
        'sentence2': "Knowledge about emerging technologies is frequently shared by experts in the online forum.",\\
        'anchor\_word': 'forum' \}\\
        
        The output should be in the following json format: \\
        \{'sentence1: S1,\\
        'sentence2': S2, \\
        'sentence3: S3,\\
        'sentence4': S4, \\
        'sentence5': S5, \\
        'anchor\_word': anchor\_word\\
        \}\\
        
        \textbf{User}: Here is the anchor word: {word}. Note that, The anchor word must appear unchanged in all the sentences.
    
\end{boxI}

%% file: prompting_box/poly_prompt.tex
\begin{boxI}\small
\label{poly_prompt}
\textit{\textbf{Polysemous Pair Generation Prompting}}: \\\
\textbf{`System Prompt'}: \\
        Using the anchor word, generate five sentences that are polysemous. Note that, the anchor word should appear in all the sentences but with different meanings. Ensure that the polysemous anchor word is positioned either in the middle or near the end of each sentence.\\
        
         Here is the example:\\
        \{ 'sentence1': "The ancient Roman forum was a bustling center of public life and political debate.",\\
        'sentence2': "The online forum became a heated battleground for discussing the latest tech trends.",\\
        'anchor\_word': 'forum' \}\\
        
         The output should be in the following json format: \\
        \{'sentence1: S1,\\
        'sentence2': S2, \\
        'sentence3: S3,\\
        'sentence4': S4, \\
        'sentence5': S5, \\
        'anchor\_word': anchor\_word \}\\
        
        \textbf{User}: Here is the anchor word: {word}.
    
\end{boxI}

%% file: prompting_box/para_prompt.tex
\begin{boxI}\small
\label{para_prompt}
\textit{\textbf{Paraphrase Task Generation Prompt}}: \\
\textbf{`System Prompt'}: \\
        Using the anchor word, create a sentence S1 that includes the anchor word. After generating S1, create four paraphrased sentences of sentence S1. All four sentences should convey the same overall meaning as S1. It's crucial that all the sentences retain the anchor word in its original form.\\
        
         For a given anchor word 'forum', the generated S1 and S2 sentences are:\\
        \{'sentence1': "The online forum provided a platform for experts to share their knowledge and engage in lively discussions about emerging technologies.",\\
        'sentence2': "A digital meeting place, the forum enabled specialists to disseminate their expertise and participate in animated conversations regarding cutting-edge innovations.",\\
        'anchor\_word': 'forum'\}\\
        
        The output should be in the following json format: \\
        \{'sentence1: S1,\\
        'sentence2': S2, \\
        'sentence3: S3,\\
        'sentence4': S4, \\
        'sentence5': S5, \\
        'anchor\_word': anchor\_word\\
        \}\\
        
        \textbf{User}: Here is the anchor word: {word}.
    
\end{boxI}

%% file: prompting_box/jumbling_prompt.tex
\begin{boxI}
\label{jumbling_box}
\textbf{Jumbling Task Data Generation}:

To create the Jumbling Task dataset, we used sentence 1 from the polysemous task dataset as the reference sentence for the Jumbling task. Next, using the reference sentence $S_1$, we generated four unique sentences by shuffling the reference sentence in four different ways:

\begin{enumerate}
    \item \textbf{$S_2$}: We first identified the location of the anchor word and then shuffled all the words present before the anchor word.
    \item \textbf{$S_3$}: We completely shuffled the entire sentence.
    \item \textbf{$S_4$} and \textbf{$S_5$}: We identified the anchor word and then exchanged one or two words around the anchor word, respectively.
\end{enumerate}
\end{boxI}

%% file: prompting_box/syn_prompt.tex
\begin{boxI}\small
\label{syn_prompt}
\textit{\textbf{Synonym Pair Generation Prompting}}: \\\
\textbf{`System Prompt'}: \\
        Using the anchor word, generate a sentence S1 of at least 15 words with the anchor word placed near the end. Next, keeping the anchor word unchanged in S2, generate a sentence S2 with the same meaning as S1 by replacing one word (other than the anchor word) with its synonym, ensuring that all word replacements occur before the anchor word in S2. \\
        
        "Note: Keep the anchor word unchanged in both sentences S1 and S2."
        Here is an example:\\
        For a given anchor word 'forum', the generated S1 and S2 sentences are:\\
        \{ 'sentence1': "Several of the questions asked by the audience in the fast-paced forum were new to the candidates.",\\
        'sentence2': "Numerous of the questions asked by the audience in the fast-paced forum were new to the candidates.",\\
        'word\_replaced': 'Several',\\
        'word\_replaced\_with': 'Numerous',\\
        'anchor\_word': 'forum' \}\\
        
        The output should be in the following json format: \\
        \{'sentence1: S1,\\
        'sentence2': S2, \\
        'word\_replaced': word, \\
        'word\_replaced\_with': new\_word, \\
        'anchor\_word': anchor\_word \}\\
        
        \textbf{User}: Follow the instructions and replace a word other than the anchor word. Here is the anchor word:\{\textbf{\textit{word}}\}. Make sure both sentences S1 and S2 have the anchor word in it."
    
\end{boxI}

%% file: prompting_box/neg_prompt.tex
\begin{boxI}\small
\label{neg_prompt}
\textit{\textbf{Negation Pair Generation Prompting}}: \\\
\textbf{`System Prompt'}: \\
        Using the anchor word, generate a sentence S1  with the anchor word in it. Next, generate a sentence S2 with an opposite meaning to S1 by adding a negation word before the anchor word in S2. Make sure the negation word is appropriate to the context of the sentence. Also, ensure that S1 and S2 should have the same words except for the negation word in S2.\\
        Note: Do not modify or change the anchor word in both sentences. \\
        
        Here is an example:
        For a given anchor word 'forum', the generated S1 and S2 sentences are:\\
        \{'sentence1': "The talented artist was adored by fans for her captivating performances.",\\
        'sentence2': "The talented artist was not adored by fans due to her underwhelming performances.",\\
        'anchor\_word': 'adored',\\
        'negation\_added': 'not adored' \}\\
        
        The output should be in the following json format: \\
        \{'sentence1: S1,\\
        'sentence2': S2, \\
        'anchor\_word': anchor\_word\\
        'negation\_added': negation\_word \}\\
        
        \textbf{User}: Here is the anchor word: {word}. 
\end{boxI}

%% file: prompting_box/anto_prompt.tex
\begin{boxI}\small
\label{anto_prompt}
\textit{\textbf{Antonym Pair Generation Prompting}}: \\\
\textbf{`System Prompt'}: \\
        Using the anchor word, generate a sentence S1 of at least 15 words with the anchor word placed near the end. Next, keeping the anchor word unchanged in S2, generate a sentence S2 with an opposite meaning to S1 by replacing one word (other than the anchor word) with its antonym, ensuring that all word replacements occur before the anchor word in S2. \\

        Note: Do not modify or change the anchor word in both sentences. \\
        Here is an example:
        For a given anchor word 'forum', the generated S1 and S2 sentences are:\\
        \{ 'sentence1':  "Several of the questions asked by the audience in the fast-paced forum were new to the candidates.",\\
        'sentence2':  "Few of the questions asked by the audience in the fast-paced forum were new to the candidates.",\\
        'word\_replaced': 'Several',\\
        'word\_replaced\_with': 'Few' \}\\
        
        The output should be in the following json format: \\
        \{'sentence1: S1,\\
        'sentence2': S2, \\
        'anchor\_word': anchor\_word\\
        'word\_replaced': word, 
        'word\_replaced\_with': new\_word  \}\\
        
        \textbf{User}: Here is the anchor word: {word}. 
\end{boxI}